\documentclass[11pt]{article}

\usepackage[preprint]{acl}
\usepackage{placeins}
\usepackage{times}
\usepackage{latexsym}
\usepackage[T1]{fontenc}
\usepackage[utf8]{inputenc}
\usepackage{microtype}
\usepackage{inconsolata}
\usepackage{graphicx}
\graphicspath{{figures/}{figures/paper/}{figures/mech/}}
 
\usepackage{amsmath,amssymb}
\usepackage{booktabs}
\usepackage{multirow}
\usepackage{xspace}
\usepackage{algorithm}
\usepackage{algpseudocode}
\usepackage{arydshln}
\usepackage[table]{xcolor}
\usepackage{stfloats} 

\definecolor{rowgrayblue}{RGB}{245,248,250}
\definecolor{rowgraygreen}{RGB}{246,249,246}
\definecolor{rowgrayyellow}{RGB}{250,249,243}
\definecolor{rowgraypink}{RGB}{250,246,247}

\newcommand{\bench}{\textsc{HopToM}\xspace}        
\newcommand{\benchfull}{\textbf{H}igher-\textbf{o}rder \textbf{p}erspective \textbf{T}heory \textbf{o}f \textbf{M}ind\xspace}
\newcommand{\ours}{\textsc{HopTrace}\xspace}

\newcommand{\V}[3]{V_{#1}[#2;\,#3]}
\newcommand{\nob}[1]{{\sim}#1}
\newcommand{\ra}{\!\rightarrow\!}

\title{Beyond Belief Nesting: \\ Rethinking Higher-Order Theory of Mind in Large Language Models}

\author{
Shanshan You$^{1,2}$, Zhengqi Wang$^{1,2}$, Hao Liu$^{3}$, Yuqin Liu$^{1,2}$, Ya He$^{1,2}$, Xin Hao$^{1,2}$, Liyuan Wang$^{1,2}$\thanks{Corresponding author.} \\
$^{1}$ Department of Psychological and Cognitive Sciences, Tsinghua University \\
$^{2}$ IDG/McGovern Institute for Brain Research, Tsinghua University \\
$^{3}$ Department of Cognitive Sciences, University of California, Irvine \\
\texttt{yss26@mails.tsinghua.edu.cn}, \texttt{liyuanwang@tsinghua.edu.cn}
  }

\begin{document}
\maketitle
\begin{abstract}
Theory of mind (ToM), the ability to reason about others' mental states, underlies effective communication and coordination. Its higher-order form enables reasoning over nested perspectives in strategic interactions. However, existing evaluations often equate belief-nesting depth with reasoning order, allowing nominally higher-order questions to admit lower-order shortcuts. We formalize effective order through the distinct events supporting successive belief prefixes, revealing widespread order inflation across existing benchmarks. We introduce \bench, with certified full-order questions and controls for lower-order alternatives. Under direct prompting, mean accuracy across four frontier LLMs falls from 73.5\% at first order to 20.0\% at fourth order, and human participants also struggle. Causal interventions link a recurring class of errors to evidence for shallower belief prefixes. We introduce \ours, a plug-and-play inference-time method that encodes each interaction once into perspective records and resolves nested queries deterministically. On certified narrative questions, it improves fourth- and fifth-order accuracy by 22.0 and 38.5 percentage points over strong prompting baselines, while using 20--39 times fewer tokens than multi-call methods. These results sharpen how higher-order ToM is evaluated and offer a practical way to improve perspective reasoning in interactive agents.
\end{abstract}


\section{Introduction}
\label{sec:intro}

\begin{figure}[!t]
\centering
\vspace{-0.2cm}
\includegraphics[width=\columnwidth]{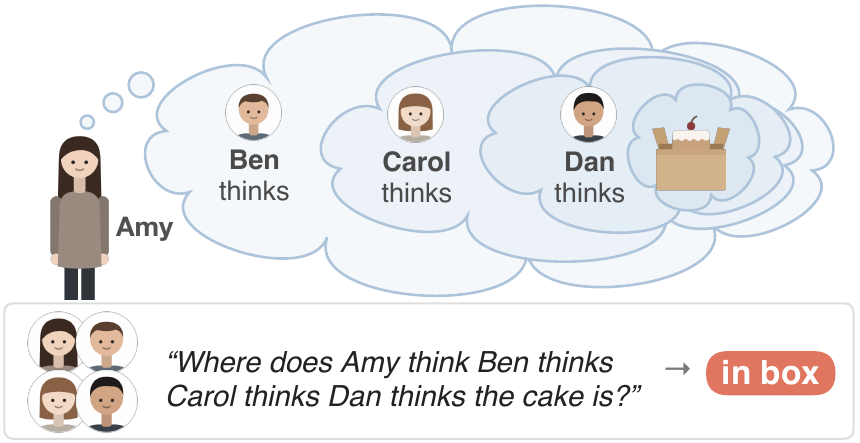}
\vspace{-0.5cm}
\caption{Nested beliefs behind a fourth-order question.}
\label{fig:teaser}
\vspace{-0.5cm}
\end{figure}

Theory of mind (ToM) allows intelligence to reason about others' beliefs, knowledge, and intentions, supporting effective communication and interaction~\citep{premack1978chimpanzee,wimmer1983beliefs}. Higher-order ToM extends this capacity to recursively nested mental states: what one agent believes another knows or believes~\citep{perner1985john}. Such reasoning becomes crucial when agents coordinate with multiple partners, information is shared selectively, and their views of one another diverge. As large language models (LLMs) increasingly participate in multi-party interactions, evaluating their ability to track these nested perspectives has become an important test of how well they can reason about others.
However, higher-order ToM is commonly evaluated by counting the belief operators in a question~\citep{kinderman1998theory,wu2023hitom,street2024llms}. Consider asking, \emph{Where does Amy \textbf{think} Ben \textbf{thinks} Carol \textbf{thinks} Dan \textbf{thinks} the cake is?} (Fig.~\ref{fig:teaser}). If all four witness the cake's last move and know that the others witnessed it, the answer follows from a single shared event. The question names four belief layers, but a respondent can answer it without distinguishing evidence for those layers. This creates a critical problem: nested wording alone \textbf{does not} establish the perspective reasoning a question tests. Analyses of human recursive mindreading likewise show that partial-chain strategies can inflate performance estimates~\citep{wilson2023recursive}.

\begin{figure*}[t]
\centering
\vspace{-0.3cm}
\includegraphics[width=0.98\textwidth]{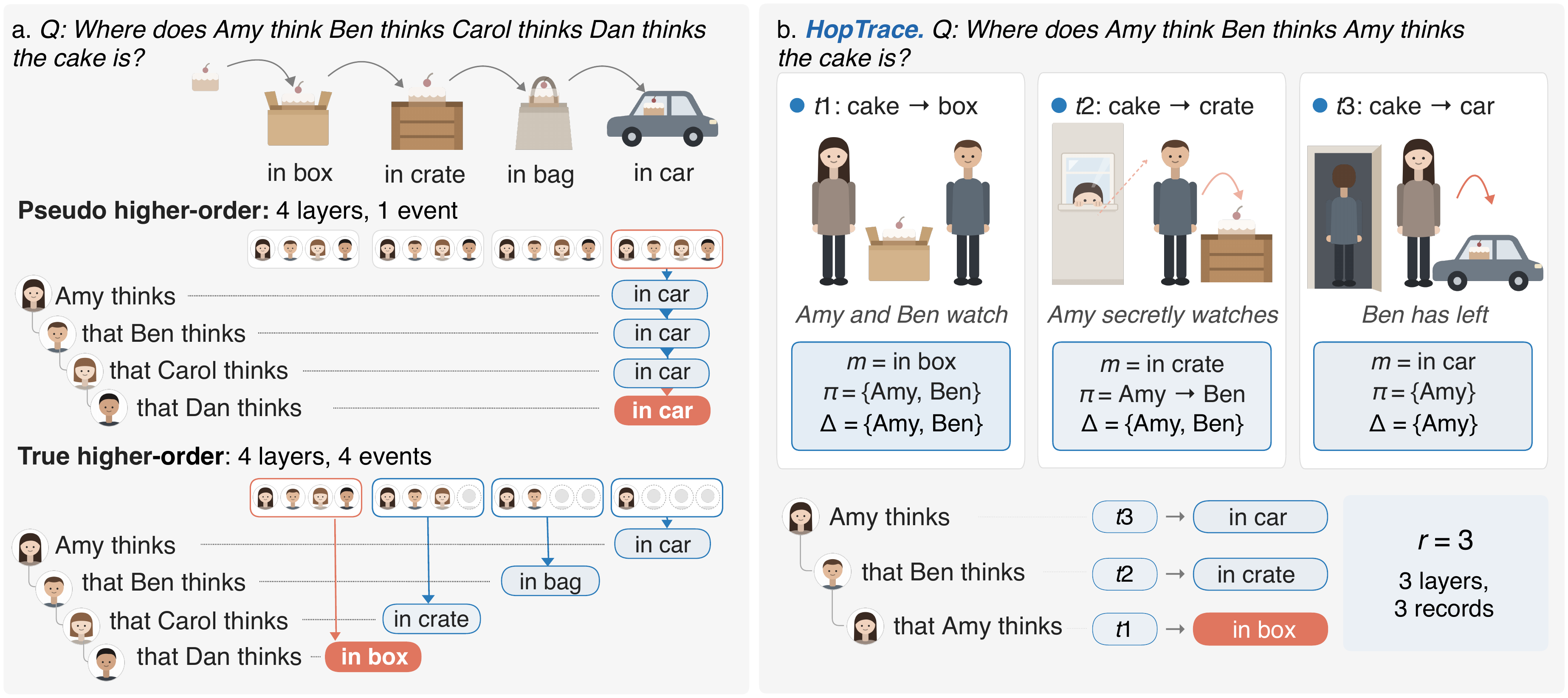}
\vspace{-0.2cm}
\caption{Belief nesting can overstate effective order.
\textbf{Left:} The same fourth-order query has $r=1$ in A and $r=4$ in B. Colors mark the event supporting each prefix.
\textbf{Right:} \ours encodes perspective records and resolves each prefix independently. Sequential backward lookup instead selects $t=2$ and incorrectly answers ``crate''.}
\label{fig:framework}
\vspace{-0.3cm}
\end{figure*}

We make this distinction precise by representing a story as belief-changing events indexed by the perspectives to which they are available. For each successive prefix of a nested query, we identify the event that settles its belief and define the query's \emph{effective order} by the number of distinct settling events. This shifts the assessment of order from the wording of a question to the information structure needed to resolve its nested perspectives. Auditing representative ToM benchmarks reveals widespread order inflation: many questions labeled as higher-order have substantially lower effective orders. Consequently, their scores alone cannot establish the depth of perspective reasoning achieved by either humans or language models.

To provide a more discriminating evaluation, we introduce \bench (\benchfull). We design the perspective structure and queries before writing each story, allowing effective order to be verified and lower-order alternatives to be examined through explicit controls. On the resulting certified questions, mean accuracy across four LLMs under direct prompting falls from 73.5\% at first order to 20.0\% at fourth. Human participants also find the certified questions difficult. Moreover, causal analyses reveal a recurring failure mode: when answering a nested question incorrectly, models often retrieve evidence supporting a shallower belief chain. These findings expose a gap between recognizing the agents mentioned in a question and maintaining the perspective relations needed to answer it.

We address this gap with \ours, a plug-and-play inference-time method that encodes an interaction once into ordered perspective records and resolves nested queries with a deterministic solver. The records preserve directed awareness, and the solver searches independently for the event supporting each belief prefix. One encoding serves multiple queries and yields an inspectable trace for each answer. On certified narrative questions, \ours improves fourth- and fifth-order accuracy by \textbf{22.0 and 38.5 percentage points} over strong prompting baselines. It uses one LLM call per interaction and \textbf{20--39 times fewer tokens} than multi-call methods. Ablation studies validate the efficacy of ordered perspectives, independent prefix search, and program execution.

Our contributions include: (1) an event-perspective formalization of effective order and an audit of order inflation in existing benchmarks; (2) a certified benchmark that verifies the intended perspective structure and includes controlled lower-order alternatives; (3) a plug-and-play inference-time method that combines reusable perspective encoding with deterministic query resolution; and (4) behavioral and causal analyses that characterize higher-order ToM performance in LLMs and explain a recurring source of model errors.

\section{Related Work}
\label{sec:related}

\paragraph{Theory of Mind (ToM).}
Psychological studies commonly assess higher-order ToM with nested-belief questions and find that performance declines as nesting deepens~\citep{kinderman1998theory,ogrady2015ease}. However, \citet{wilson2023recursive} showed that some deeply nested items admit partial-chain solutions, with performance dropping sharply when those shortcuts are blocked. Epistemic logic formally represents nested beliefs and changes in agents' information~\citep{fagin1995reasoning,baltag1998logic}. These lines of work highlight a distinction between the depth of a question and the belief layers its story makes distinguishable. Our effective-order formulation measures this distinction at the item level through the events supporting successive belief prefixes.

\paragraph{ToM in Language Models.}
LLM evaluations have expanded from classic false-belief tasks to dialogues and deeper belief questions~\citep{kosinski2024evaluating,kim2023fantom,wu2023hitom,street2024llms}. Existing efforts have likewise used perspective filtering, temporal belief tracking, and recursive prompting to improve ToM performance~\citep{wilf2023think,hou2024timetom,lei2026mind}, while mechanistic work has identified retrieval operations involved in belief tracking~\citep{prakash2025lookbacks}. Yet nesting-depth labels do not reveal how many distinct events support a query's belief layers, leaving the source of success on higher-order questions unclear. We use effective order to audit this gap, design controlled evaluations, and guide inference over explicit perspective records.

\section{True Higher-Order ToM}
\label{sec:problem}

Belief-nesting depth can misclassify shortcut-solvable questions as higher-order. We define \emph{effective order} from the distinct events supporting successive belief prefixes, giving a structural criterion for genuinely higher-order questions, and then perform an in-depth empirical analysis.

\subsection{Problem Formulation}
\label{sec:formulation}

\paragraph{Interactions and queries.}
Consider human or artificial agents interacting over time. Their observations and communications may change both their beliefs about a target variable $x$, such as an object's location, and their beliefs about one another's beliefs. A higher-order ToM query is a belief chain $q=[q_1,\ldots,q_k]$, read from the outside in. For example, $[a,c,b]$ asks what $a$ believes $c$ believes $b$ believes about $x$. Its \emph{syntactic order} is $k=|q|$.

\paragraph{Perspective records.}
We encode the interaction as time-ordered records $E_t=\V{t}{\pi_t}{\Delta_t}(m_t)$. Each record carries a value $m_t$ of $x$, which may be initially held, observed, or communicated and need not be correct. The structure $\pi_t$ specifies agents' access to this value and their beliefs about others' access. The set $\Delta_t$ lists agents whose own belief becomes $m_t$ and may be empty when only beliefs about others change. We write $\pi_t$ as a chain of groups $G_1\ra\cdots\ra G_\ell$. Access is common knowledge within a group, while an arrow denotes one-sided observation of the next group's access. A pseudo-observer $\nob{c}$ is an agent whom others incorrectly believe to have received $m_t$. These records cover observation, communication, deception, and misperception. Initial states are encoded to support every queried prefix, ensuring that a latest supporting record exists. Beliefs persist until updated.

\paragraph{Effective order.}
Let $p_j=[q_1,\ldots,q_j]$ be the $j$-th prefix of $q$. A record \emph{supports} $p_1$ if $q_1\in\Delta_t$. For $j\geq2$, it supports $p_j$ if $\pi_t$ covers the chain. This requires $q_1$ to be an actual observer in the outermost group. Each successive agent must be represented, actually or presumptively, in the same group or the next group along a directed observation link. For example, $\{a,b\}\ra c$ covers $[a,c]$ and $[a,b,c]$, but not $[c,a]$.
We resolve each prefix independently against the full interaction history. Let $s_j$ be the index of the latest record supporting $p_j$. That record determines the value of $x$ attributed under $p_j$. The record indexed by $s_k$ determines the answer to the full query. The \emph{effective order} is
\begin{equation}
r(q,x)=\bigl|\{s_1,\ldots,s_k\}\bigr|\leq k.
\end{equation}
A query has \emph{full effective order} if $r(q,x)=k$, meaning that its successive prefixes have distinct latest supporting events. This rules out shared-event collapse. Coinciding answers to lower-order queries are assessed separately.

\begin{table}[t]
\centering
\vspace{-0.1cm}
\footnotesize
\setlength{\tabcolsep}{3pt}
\resizebox{0.9\columnwidth}{!}{
\begin{tabular}{@{}l r c c c c@{}}
\toprule
\textbf{Benchmark} & $N$ & $k_0$ & $k$ & \textbf{Mean $r$} & \textbf{Full order} \\
\midrule
ToMi       & 1,998 & 2 & 2       & 1.36 & 36.2\% \\
ExploreToM & 3,298 & 2 & 2       & 1.07 & 6.6\%  \\
Hi-ToM     & 120   & 4 & 4       & 2.07 & 2.5\%  \\
\addlinespace[0.3ex]
\cdashline{1-6}[2pt/2pt]
\addlinespace[0.8ex]
\bench (Ours)    & 200   & 5 & 5    & 5.00 & 100\%  \\
\bottomrule
\end{tabular}
}
\vspace{-0.1cm}
\caption{Effective order at each benchmark's highest published level. $N$ is the number of questions, $k$ is the belief-chain length. Full order is the share with $r=k$.}
\label{tab:audit}
\end{table}
\subsection{Rethinking Previous ToM Benchmarks}

\label{sec:audit}

We audit questions at the highest published level in three existing benchmarks: ToMi~\citep{le2019revisiting}, ExploreToM~\citep{sclar2025explore}, and Hi-ToM~\citep{wu2023hitom}. For these benchmarks, we derive perspective records from their generators' action logs. The audit is independent of released answer labels. Appendix~\ref{app:audit} reports encoding checks, label inconsistencies, and results at other orders.

The published higher-order labels substantially overstate the order tested (Table~\ref{tab:audit}). At the highest level of each benchmark, only 36.2\% of ToMi questions, 6.6\% of ExploreToM questions, and 2.5\% of Hi-ToM questions have full effective order.
Hi-ToM shows how this inflation grows with nesting depth (Table~\ref{tab:hitom_audit}). From second to fourth order, mean effective order rises only from 1.40 to 2.07, while the full-order share falls from 40.0\% to 2.5\%. At fourth order, 89 of 120 questions have $r\leq2$, and only 6.7\% of those with claim events reach full order. An additional shortcut arises from answer coincidence: the correct answer matches the answer to a shorter prefix or the target's actual final location in 66.7\% of second-order and 85.0\% of fourth-order questions. 

Three patterns produce structural collapse. In \emph{subsumed perspectives}, an added agent shares the event supporting an outer layer (Fig.~\ref{fig:framework}, left). In \emph{idle gaps}, agents have different observation histories, but the intervening events do not change $x$. In \emph{reported mental states}, a single claim or explicit statement settles multiple prefixes at once. Subsumed perspectives, idle gaps, and reported mental states account for 54\%, 33\%, and 13\% of collapsed layers in Hi-ToM, respectively. The audited benchmarks are therefore dominated by \emph{pseudo-higher-order} items: their questions contain nested beliefs, but their interactions do not require a distinct supporting event for each layer. Establishing higher-order ToM requires certifying these perspective distinctions and controlling shortcuts in the answers.

\begin{table}[t]
\centering
\vspace{-0.1cm}
\small
\setlength{\tabcolsep}{2pt}
\resizebox{0.9\columnwidth}{!}{
\begin{tabular}{@{}l r c c c c@{}}
\toprule
\textbf{Subset} & $k$ & \textbf{Mean $r$} & \textbf{Full order} & \textbf{State match} & \textbf{$r$ counts} \\
\midrule
\multirow{4}{*}{All} & 1 & 1.00 & 100\% & 50.0\% & 120 \\
    & 2 & 1.40 & 40.0\% & 66.7\% & 72 / 48 \\
    & 3 & 1.68 & 9.2\% & 79.2\% & 50 / 59 / 11 \\
    & 4 & 2.07 & 2.5\% & 85.0\% & 26 / 63 / 28 / 3 \\
\midrule
\multirow{3}{*}{w/o Claims} & 2 & 1.29 & 29.3\% & 73.3\% & 53 / 22 \\
    & 3 & 1.47 & 2.7\% & 86.7\% & 42 / 31 / 2 \\
    & 4 & 1.84 & 0.0\% & 89.3\% & 24 / 39 / 12 / 0 \\
\midrule
\multirow{3}{*}{w/ Claims} & 2 & 1.58 & 57.8\% & 55.6\% & 19 / 26 \\
    & 3 & 2.02 & 20.0\% & 66.7\% & 8 / 28 / 9 \\
    & 4 & 2.44 & 6.7\% & 77.8\% & 2 / 24 / 16 / 3 \\
\bottomrule
\end{tabular}
}
\vspace{-0.1cm}
\caption{Hi-ToM audit by belief-chain length. Each order contains 120 unique questions, including 75 without and 45 with claim events. Full order is the share with $r=k$. State match is the share whose answer equals the answer to a shorter prefix or the target's actual final location. The last column gives counts for $r=1,\ldots,k$.}
\vspace{-0.5cm}
\label{tab:hitom_audit}
\end{table}

\subsection{\bench: A Certified Benchmark for Higher-Order ToM}
\label{sec:bench}

\paragraph{Perspective-first construction.}
The audit motivates specifying the perspective structure and queries before writing their natural-language realizations. For each query, we identify the record supporting every belief prefix and certify \emph{full effective order} when $r=k$. In our certified items, these supporting records also occur in decreasing time order from the outermost to the innermost prefix. This temporal constraint is part of the benchmark design, not the definition of effective order. We vary the positions of supporting events and interleave other updates to reduce the regularity of simple backward-stepping heuristics.

\paragraph{Scenarios and questions.}
We design 10 prototypes spanning 10 scenario types in six families, including asymmetric observation, deception, misperception, and delayed information (Table~\ref{tab:scenarios}). Each prototype yields 10 variants that preserve its perspective structure while changing names, settings, objects, and wording. Each variant is realized as both a narrative and a multi-party dialogue. The resulting benchmark contains 200 story realizations of 100 underlying interactions.

Each realization has five certified questions, one at each syntactic order from one to five, and four controls targeting collapsed belief layers or coinciding lower-order answers. This yields 1,000 certified questions and 800 controls. Questions use open-ended answers to avoid reliance on candidate options~\citep{balepur2024artifacts,zheng2024robust}. We also provide a multiple-choice version (Appendix~\ref{app:mc}). All realizations were checked for fluency, and each certified question was manually verified against its supporting-record trace.

\subsection{Empirical Analysis}
\label{sec:empirical}

\begin{table}[t]
\centering
\footnotesize
\setlength{\tabcolsep}{4pt}
\begin{tabular}{@{}l ccccc c@{}}
\toprule
\textbf{Model} & \textbf{1st} & \textbf{2nd} & \textbf{3rd} & \textcolor{blue}{\textbf{4th}} & \textcolor{blue}{\textbf{5th}} & \textcolor{red}{\textbf{All}} \\
\midrule
GPT-6 Sol       & 92.0 & 62.0 & 58.5 & 32.5 & 36.0 & 56.2 \\
Claude Opus 5.5 & 69.5 & 33.0 & 42.5 & 16.0 & 21.5 & 36.5 \\
Gemini 3 Flash  & 72.5 & 34.5 & 41.0 & 17.0 & 16.0 & 36.2 \\
DeepSeek-V4-Pro & 60.0 & 49.0 & 32.5 & 14.5 & 14.0 & 34.0 \\
\midrule
Average & 73.5 & 44.6 & 43.6 & 20.0 & 21.9 & 40.7 \\
\bottomrule
\end{tabular}
\caption{Direct-prompting accuracy (\%) on the 1,000 certified full-order questions of \bench. Each order contains 200 questions.}
\vspace{-0.2cm}
\label{tab:direct}
\end{table}

\begin{figure}[th]
\centering
\includegraphics[width=\columnwidth]{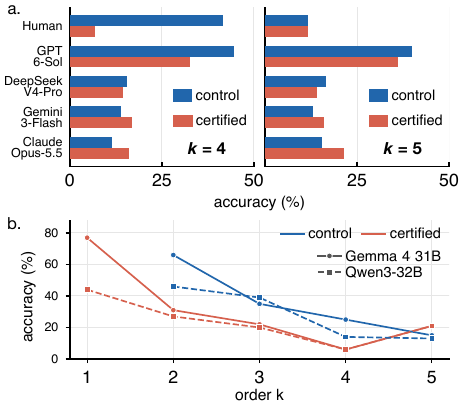}
\vspace{-0.5cm}
\caption{Results on certified questions and controls. (a) Fourth- and fifth-order results for human participants, who could not reread the stories, and four frontier LLMs. (b) Results by syntactic order for open-weight models.}
\vspace{-0.2cm}
\label{fig:truepseudo}
\end{figure}

We evaluate four frontier LLMs under direct prompting and two open-weight models, Gemma 4 31B and Qwen3-32B. A complementary human pilot uses three stories that participants could not reread while answering (Appendix~\ref{app:human}).

\paragraph{Certified questions become difficult at higher orders.}
Across the four frontier LLMs, mean accuracy falls from 73.5\% at first order to 20.0\% at fourth order (Table~\ref{tab:direct}). GPT-6 Sol drops from 92.0\% to 32.5\%, and the other three score 14.5--17.0\% at fourth order. Fifth-order accuracy stays low, averaging 21.9\%. Strong performance on simpler belief questions therefore does not carry over to certified fourth- and fifth-order questions under direct prompting. Human participants also struggle, with 33.3\% accuracy on certified questions overall and about 9\% at orders four and five combined (Fig.~\ref{fig:truepseudo}a). The slight rise from fourth to fifth order for GPT-6 Sol and Claude Opus 5.5 may reflect how certified items are built. Here, supporting events run backward in time across prefixes, so the deepest layer falls on one of the story's earliest belief changes. A model that names an early location may then answer correctly.

\paragraph{Controls reveal when lower-order alternatives help.}
At fourth order, human accuracy is 41.7\% on controls but only 6.7\% on certified questions (Fig.~\ref{fig:truepseudo}a). GPT-6 Sol shows the same direction of contrast, though less strongly. The other frontier LLMs perform poorly in both conditions. Gemma 4 31B and Qwen3-32B show clear control advantages at orders two through four (Fig.~\ref{fig:truepseudo}b). At fifth order, human accuracy is low in both conditions, and the control advantage narrows or reverses for the weaker models. These patterns show that lower-order alternatives can substantially inflate performance when respondents can exploit them. When both conditions are difficult, their accuracy gap becomes less informative about the underlying difference in perspective demands.




\section{Method}
\label{sec:method}

Full-order questions require tracking distinct perspective updates and selecting the evidence that supports each belief prefix. We introduce \ours, a plug-and-play inference-time method that makes these lookups explicit while reusing one LLM encoding across multiple queries.

\paragraph{Overview.}
Let $S$ denote an account of an interaction, such as a narrative, dialogue, or event log. Our formulation associates each prefix of a belief-chain query with its latest supporting record (Sec.~\ref{sec:formulation}). Interpreting the same interaction afresh for every query repeats the work of tracking agents' perspectives, while chaining partial answers can select the wrong records. \ours separates perspective encoding from query resolution (Fig.~\ref{fig:framework}, right). An LLM encodes $S$ once, and a deterministic solver answers queries from the resulting records:
\begin{align}
\mathcal{V}
&=\operatorname{Enc}(S)
 =\langle E_1,\ldots,E_n\rangle,
 \label{eq:enc}\\
(\hat y,\tau)
&=\operatorname{Solve}(\mathcal{V},q,x),
 \quad \hat y=m_{s_k}.
 \label{eq:solve}
\end{align}
Here $E_t=\V{t}{\pi_t}{\Delta_t}(m_t)$ is a perspective record tagged with its target variable, which is omitted from the notation when $x$ is fixed. The trace $\tau=\langle(p_j,E_{s_j})\rangle_{j=1}^k$ identifies the supporting record for every prefix.

\paragraph{D1: Ordered perspectives.}
Who observes an event and who knows about that observation can lead to different nested beliefs. The encoder therefore preserves joint access, one-sided awareness, and presumed observation in the ordered perspective structure $\pi_t$. It records changes to agents' own beliefs in $\Delta_t$. A flat set of agents cannot represent these distinctions.

\paragraph{D2: Independent per-prefix search.}
The record supporting a prefix depends on that prefix's perspective, not selected for the preceding prefix. Stepping backward through individual agents' first-order beliefs can therefore select the wrong evidence, even when the correct supporting records are ordered in time. The solver instead searches the full record sequence independently for each $p_j$.

\paragraph{D3: Program execution.}
Applying perspective coverage and temporal selection across several prefixes can itself be a source of model error. We implement these rules as the deterministic procedure in Algorithm~\ref{alg:backtrack}. The solver uses only the encoded records and returns an inspectable trace with each answer. It is exact relative to those records under the specified support rule. Encoding requires one LLM call per $S$, after which each query takes at most $k$ backward scans of the relevant records.

\begin{algorithm}[t]
\small
\caption{Deterministic query resolution}
\label{alg:backtrack}
\begin{algorithmic}[1]
\Require Time-ordered records $\mathcal{V}=\langle E_1,\ldots,E_n\rangle$,
query $q=[q_1,\ldots,q_k]$, target $x$
\State $\mathcal{V}_x\gets$ records tagged with $x$, retaining their indices in $\mathcal{V}$
\State $\tau\gets[\,]$
\For{$j=1$ \textbf{to} $k$}
    \State $p_j\gets[q_1,\ldots,q_j]$
    \For{$E_t$ in $\mathcal{V}_x$ from latest to earliest}
        \If{$E_t$ supports $p_j$}
            \State $s_j\gets t$
            \State Append $(p_j,E_t)$ to $\tau$
            \State \textbf{break}
        \EndIf
    \EndFor
\EndFor
\State $\hat y\gets m_{s_k}$ \Comment{value carried by $E_{s_k}$}
\State \Return $(\hat y,\tau)$
\end{algorithmic}
\end{algorithm}

\section{Experiments}
\label{sec:experiments}

In this section, we first describe the experimental setup and then analyze main results. 

\subsection{Experimental Setup}
\label{sec:setup}

\paragraph{Evaluation benchmarks.}
\bench contains 100 underlying interactions, each realized as a narrative and a dialogue, for 200 texts and 1,800 question instances. The primary method comparison uses the 100 narratives and their 500 certified full-order questions, with 100 questions at each order. Their 400 controls are used in the ablation analysis. The evaluation in Sec.~\ref{sec:empirical} uses both language formats.
We also evaluate on Hi-ToM~\citep{wu2023hitom}, which has 120 stories and 600 unique questions spanning orders zero to four. Each question is released in two prompt formats. Dataset handling is detailed in Appendix~\ref{app:audit}.

\paragraph{Models and baselines.}
We evaluate GPT-6 Sol~\citep{openai2026gpt6}, Claude Opus 5.5~\citep{anthropic2026claude}, Gemini 3 Flash~\citep{google2026gemini3}, and DeepSeek-V4-Pro~\citep{deepseek2026v4}. Gemma 4 31B~\citep{google2026gemma4} and Qwen3-32B~\citep{yang2025qwen3} serve as open-weight encoders. We compare \ours\ with direct answering, chain-of-thought prompting (CoT)~\citep{wei2022chain}, TimeToM~\citep{hou2024timetom}, and RecToM~\citep{lei2026mind}. Comparisons within each model use the same underlying LLM. Prompt adaptations and execution settings are described in Appendix~\ref{app:time}.

\begin{table}[!t]
\centering
\footnotesize
\setlength{\tabcolsep}{2.8pt}
\begin{tabular}{@{}c l ccccc c c@{}}
\toprule
& & \multicolumn{6}{c}{\textbf{\bench (True Higher-Order)}} & \textbf{Hi-ToM} \\
\cmidrule(lr){3-8}\cmidrule(lr){9-9}
& \textbf{Method} & \textbf{1st} & \textbf{2nd} & \textbf{3rd} & \textcolor{blue}{\textbf{4th}} & \textcolor{blue}{\textbf{5th}} & \textcolor{red}{\textbf{All}} & \textcolor{red}{\textbf{All}} \\
\midrule
\multirow{4}{*}{\rotatebox{90}{\scriptsize GPT-6 Sol}}
& CoT     & \textbf{94} & 58 & 52 & 42 & 44 & 58.0 & 80.3 \\
& TimeToM & 78 & \textbf{83} & 56 & 58 & 50 & 65.0 & 78.5 \\
& RecToM  & 85 & 74 & 52 & 58 & 36 & 61.0 & 79.2 \\
\addlinespace[0.3ex]
\cdashline{2-9}[2pt/2pt]
\addlinespace[0.8ex]
& \ours   & 89 & 70 & \textbf{67} & \textbf{76} & \textbf{90} & \textbf{78.4} & \textbf{82.4} \\
\midrule
\multirow{4}{*}{\rotatebox{90}{\scriptsize DeepSeek}}
& CoT     & 76 & 68 & 35 & 23 & 14 & 43.2 & 62.5 \\
& TimeToM & 55 & 48 & 44 & 53 & 41 & 48.2 & 65.8 \\
& RecToM  & 62 & 49 & 43 & 47 & 38 & 47.8 & 61.3 \\
\addlinespace[0.3ex]
\cdashline{2-9}[2pt/2pt]
\addlinespace[0.8ex]
& \ours   & \textbf{94} & \textbf{69} & \textbf{71} & \textbf{70} & \textbf{81} & \textbf{77.0} & \textbf{87.6} \\
\midrule
\multirow{4}{*}{\rotatebox{90}{\scriptsize Gemini-Flash}}
& CoT     & \textbf{98} & 63 & 56 & 43 & 40 & 60.0 & 70.6 \\
& TimeToM & 79 & \textbf{64} & 38 & 51 & 39 & 54.2 & 45.6 \\
& RecToM  & 74 & 61 & 35 & 16 & 42 & 45.6 & 67.5 \\
\addlinespace[0.3ex]
\cdashline{2-9}[2pt/2pt]
\addlinespace[0.8ex]
& \ours   & 91 & \textbf{64} & \textbf{68} & \textbf{73} & \textbf{74} & \textbf{74.0} & \textbf{87.6} \\
\midrule
\multirow{4}{*}{\rotatebox{90}{\scriptsize Claude-Opus}}
& CoT     & \textbf{93} & 59 & 56 & 49 & 41 & 59.6 & 79.8 \\
& TimeToM & 78 & \textbf{75} & 45 & 31 & 40 & 53.8 & 80.0 \\
& RecToM  & 77 & \textbf{75} & 66 & 59 & 49 & 65.2 & 79.8 \\
\addlinespace[0.3ex]
\cdashline{2-9}[2pt/2pt]
\addlinespace[0.8ex]
& \ours   & \textbf{93} & 65 & \textbf{85} & \textbf{62} & \textbf{79} & \textbf{76.8} & \textbf{81.9} \\
\midrule
\multirow{4}{*}{\rotatebox{90}{\scriptsize Average}}
& CoT     & 90.3 & 62.0 & 49.8 & 39.3 & 34.8 & 55.2 & 73.3 \\
& TimeToM & 72.5 & \textbf{67.5} & 45.8 & 48.3 & 42.5 & 55.3 & 67.5 \\
& RecToM  & 74.5 & 64.8 & 49.0 & 45.0 & 41.3 & 54.9 & 72.0 \\
\addlinespace[0.3ex]
\cdashline{2-9}[2pt/2pt]
\addlinespace[0.8ex]
& \ours   & \textbf{91.8} & 67.0 & \textbf{72.8} & \textbf{70.3} & \textbf{81.0} & \textbf{76.6} & \textbf{84.9} \\
\bottomrule
\end{tabular}
\caption{Accuracy (\%) on \bench and Hi-ToM. Models: GPT-6 Sol, DeepSeek-V4-Pro, Gemini 3 Flash, and Claude Opus 5.5. Best per model and column in bold.}
\vspace{-0.2cm}
\label{tab:main}
\end{table}

\subsection{Experimental Results}
\label{sec:results}

\begin{figure*}[t]
\centering
\vspace{-0.2cm}
\includegraphics[width=\textwidth]{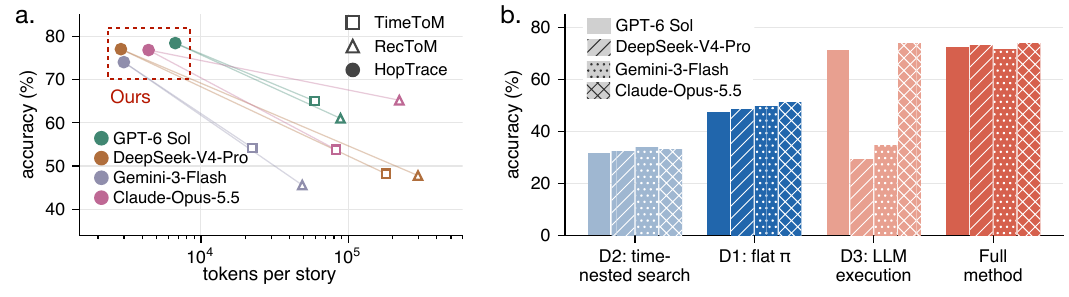}
\vspace{-0.7cm}
\caption{Accuracy, inference cost, and ablations on the narrative part of \bench. (a) Accuracy on 500 certified full-order questions against tokens per story. \ours answers all nine questions per story with one encoding call, while TimeToM and RecToM answer the five certified questions. (b) Accuracy of the full method and its ablations on all 900 narrative questions. Hatching denotes individual models and dashed lines denote model averages.}
\vspace{-0.2cm}
\label{fig:results2}
\end{figure*}

\paragraph{Overall performance.}
On the 500 certified narrative questions, \ours reaches 76.6\% mean accuracy across four LLMs (Table~\ref{tab:main}). CoT, TimeToM, and RecToM reach 55.2\%, 55.3\%, and 54.9\%, respectively, giving \ours an advantage of \textbf{at least 21.3 percentage points}. The difference is largest at \textbf{higher orders}. At fourth and fifth order, \ours reaches \textbf{70.3\% and 81.0\%}, compared with 48.3\% and 42.5\% for the strongest baseline at each order.
Our method also improves open-weight models on all 1,800 questions. Gemma 4 31B rises from 33.2\% under direct prompting to 67.7\% with \ours, while Qwen3-32B rises from 25.3\% to 38.7\%. Qwen's smaller gain coincides with encodings that often omit directed or presumed observations. Since the solver is unchanged across encoders, the quality of the perspective records remains consequential.

\paragraph{Efficiency.}
\ours uses one LLM call and 4,235 tokens per story on average while answering all nine questions (Fig.~\ref{fig:results2}a). TimeToM uses 34.1 calls and 86,259 tokens per story, and RecToM uses 32.4 calls and 164,641 tokens, each for five certified questions. Even if the full encoding cost of \ours is assigned to those five questions, its token use is \textbf{20.4$\times$ and 38.9$\times$ lower}, respectively. Wall-clock measurements show the same pattern, with 4--217\,s per story for \ours and 268--1,258\,s for the two recursive baselines across models (Appendix~\ref{app:impl}).

\begin{figure}[t]
\vspace{-0.1cm}
\centering
\includegraphics[width=\columnwidth]{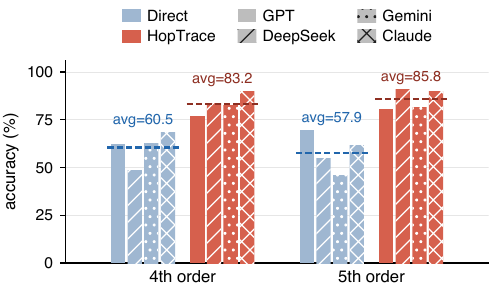}
\vspace{-0.5cm}
\caption{Fourth- and fifth-order accuracy on the certified questions of \textsc{\ours} with reasoning enabled.}
\vspace{-0.3cm}
\label{fig:thinking}
\end{figure}

\paragraph{Reasoning models.}
Extended thinking strengthens direct answering, providing a more demanding test of \ours. With thinking enabled for both direct answering and perspective encoding, \ours raises mean accuracy from 60.5\% to 83.2\% at fourth order and from 57.9\% to 85.8\% at fifth order on \bench (Fig.~\ref{fig:thinking}). All four models improve at both orders. The gains are most pronounced on the higher-order questions that \ours targets, showing that explicit perspective records and deterministic prefix resolution remain valuable even for reasoning-enabled models. As one encoding serves multiple queries, these gains do not require a separate thinking-enabled LLM call for each question.

\paragraph{Transfer to Hi-ToM.}
\label{sec:generalization}
\ours also achieves the highest accuracy for every model on Hi-ToM, ranging from 81.9\% to 87.6\% (Table~\ref{tab:main}). This shows that the encoder--solver pipeline applies to a separately constructed benchmark with different story conventions and a multiple-choice answer format. The result has a narrower interpretation for higher-order ability: as the audit in Sec.~\ref{sec:audit} shows, only 2.5\% of Hi-ToM's nominally fourth-order questions have full effective order. Aggregate Hi-ToM accuracy therefore provides limited evidence about performance on full-order questions.

\paragraph{Ablation study.}
\label{sec:ablation}
The three ablations correspond to D1--D3 in Sec.~\ref{sec:method} (Fig.~\ref{fig:results2}b). Flattening the ordered perspectives in D1 reduces average accuracy by 23.5 points. The loss is consistent across models, showing the value of distinguishing joint access from directed and presumed observation. Replacing independent prefix search with time-nested first-order lookup in D2 causes the largest drop, 39.9 points. Chronology alone does not identify which event supports a nested perspective. In D3, LLM execution instead of program execution costs 20.4 points on average. GPT-6 Sol and Claude Opus 5.5 remain close to the full method, while DeepSeek-V4-Pro and Gemini 3 Flash fall to 29.2\% and 34.4\%. Deterministic execution thus makes the encoded records substantially more useful across models.
On the 90 prototype questions, replacing model-generated records with gold records raises accuracy from 73.1\% to 100\%. The solver can resolve these questions from correctly specified records, leaving perspective encoding as the source of the observed gap on this subset.

\begin{figure}[t]
\centering
\includegraphics[width=\columnwidth]{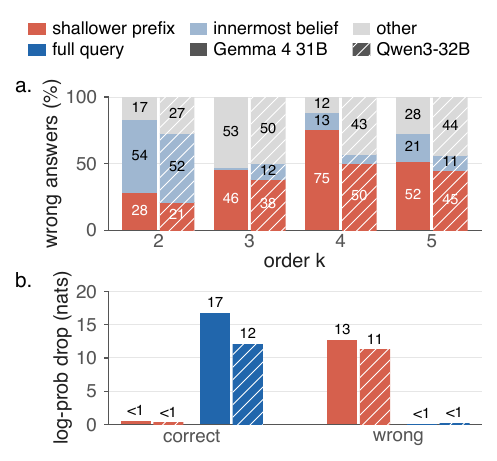}
\vspace{-0.6cm}
\caption{Direct answers. (a) Wrong answers by order $k$ (\%). \emph{Shallower prefix}: answer to a shorter prefix. \emph{Innermost belief}: the last-named agent's belief. (b) Log-probability drop when supporting paragraph is hidden.
}
\label{fig:mech-diag}
\vspace{-0.2cm}
\end{figure}

\begin{figure}[t]
\centering
\includegraphics[width=\columnwidth]{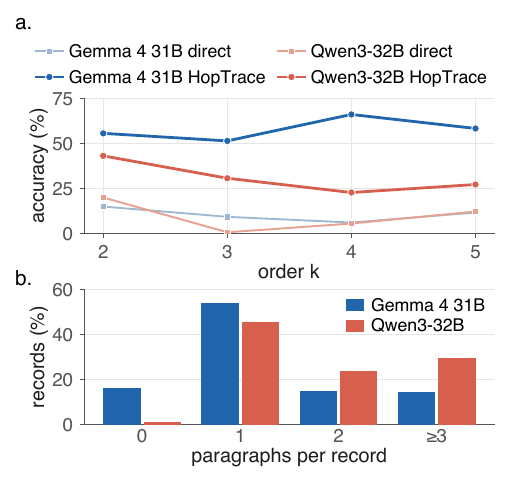}
\vspace{-0.6cm}
\caption{\ours vs.\ direct answering. (a) Accuracy on shortcut-resistant certified questions. (b) Paragraphs each record depends on.}
\label{fig:mech-hoptrace}
\vspace{-0.2cm}
\end{figure}

\paragraph{Mechanism.}
\label{sec:mechanism}
We use attention interventions and activation patching to examine direct answers from Gemma 4 31B and Qwen3-32B (Appendix~\ref{app:mech}). Many wrong answers repeat the answer to a shallower prefix of the query, including 75\% for Gemma and 50\% for Qwen3 at fourth order (Fig.~\ref{fig:mech-diag}a). At second order, most errors give the innermost agent's own belief. For correct answers, blocking the paragraph containing the full query's supporting event reduces answer log-probability by 12--17 nats, while blocking intermediate-prefix paragraphs has much smaller effects (Fig.~\ref{fig:mech-diag}b). Shallower-prefix errors reverse this pattern: that prefix's paragraph has the largest effect in 81\% and 90\% of cases, and the full query's paragraph has little effect. These errors come from evidence selected for the wrong perspective.

With \ours, the model only encodes the story into perspective records, and the solver resolves the query prefix by prefix. On shortcut-resistant certified questions, accuracy rises from about 10\% under direct answering to 58\% with Gemma and 31\% with Qwen3 (Fig.~\ref{fig:mech-hoptrace}a). Encoding is a local task: about half of the records draw on a single story paragraph (54\% for Gemma and 46\% for Qwen3; Fig.~\ref{fig:mech-hoptrace}b). Patching shows that each agent's information moves to the final token directly from its name, not through the other agents (Fig.~\ref{fig:mechA3}). For correct answers, the record selected by the solver draws on the full query's supporting paragraph in 89\% of cases for both models. Direct answering thus tends to stop short when tracing the chain back, whereas \ours leaves this backtracking to the solver and the model to the local encoding it handles well.

\section{Conclusion and Discussion}
\label{sec:conclusion}

Higher-order ToM cannot be assessed reliably from the nesting depth of a question alone. We formalize the perspective distinctions created by an interaction through effective order, reveal their frequent collapse in existing benchmarks, and construct a controlled evaluation that exposes substantial difficulty for both LLMs and humans. We also show that explicit perspective records, combined with deterministic query resolution, markedly improve model performance. Together, these findings provide a clearer basis for measuring higher-order ToM and a practical direction for strengthening perspective reasoning as interactions unfold.

\paragraph{Perspective tracking as a bottleneck.}
Models often retrieve evidence appropriate for a shallower prefix, whereas the solver answers all prototype questions given gold records. This points to perspective encoding and evidence selection as major constraints in the evaluated systems. Human recursive mindreading is also sensitive to task demands~\citep{stiller2007perspective,wilson2023recursive}. Whether explicit records help human participants remains an open question.

\paragraph{Writing higher-order questions.}
Effective order can be checked before an interaction is written in natural language. Full effective order excludes shared-event collapse, but neither rules out coinciding answers nor establishes the minimum computation a respondent uses. Future benchmarks should report effective alongside syntactic order, include lower-order answer controls, and vary event timing to limit positional shortcuts.

\paragraph{Dividing the labor.}
One perspective encoding serves multiple queries, while program execution produces an inspectable supporting trace. This division suggests a way to reason over longer interactions, where agents must continually update what they believe others have seen, learned, or communicated. Improving the fidelity of those records remains essential.

\clearpage
\section*{Limitations}
\label{sec:limitations}
Our formalism focuses on discrete beliefs about specified target variables, leaving uncertainty, graded trust, and beliefs inferred from indirect cues for future extensions. \bench uses controlled written interactions, which could be extended to longer exchanges and multimodal observations. Our mechanistic interventions examine two open-weight models, and similar analyses across other model families would broaden their scope.

\section*{Acknowledgments}
This work was supported by the NSFC Project (No.~T2622023), the Beijing Nova Program (No.~202604841279), and the Beijing Major Science and Technology Project (No.~Z251100008425003).


\bibliography{custom}

@article{premack1978chimpanzee, 
title={Does the chimpanzee have a theory of mind?}, 
volume={1}, 
DOI={10.1017/S0140525X00076512}, 
number={4}, 
journal={Behavioral and Brain Sciences}, 
author={Premack, David and Woodruff, Guy}, 
year={1978}, 
pages={515–526}}

@article{wimmer1983beliefs,
title = {Beliefs about beliefs: Representation and constraining function of wrong beliefs in young children's understanding of deception},
journal = {Cognition},
volume = {13},
number = {1},
pages = {103-128},
year = {1983},
issn = {0010-0277},
doi = {https://doi.org/10.1016/0010-0277(83)90004-5},
url = {https://www.sciencedirect.com/science/article/pii/0010027783900045},
author = {Heinz Wimmer and Josef Perner}
}

@article{perner1985john,
title = {“John thinks that Mary thinks that…” attribution of second-order beliefs by 5- to 10-year-old children},
journal = {Journal of Experimental Child Psychology},
volume = {39},
number = {3},
pages = {437-471},
year = {1985},
issn = {0022-0965},
doi = {https://doi.org/10.1016/0022-0965(85)90051-7},
url = {https://www.sciencedirect.com/science/article/pii/0022096585900517},
author = {Josef Perner and Heinz Wimmer}
}

@article{kinderman1998theory,
author = {Kinderman, Peter and Dunbar, Robin and Bentall, Richard P.},
title = {Theory-of-mind deficits and causal attributions},
journal = {British Journal of Psychology},
volume = {89},
number = {2},
pages = {191-204},
doi = {https://doi.org/10.1111/j.2044-8295.1998.tb02680.x},
url = {https://bpspsychub.onlinelibrary.wiley.com/doi/abs/10.1111/j.2044-8295.1998.tb02680.x},
eprint = {https://bpspsychub.onlinelibrary.wiley.com/doi/pdf/10.1111/j.2044-8295.1998.tb02680.x},
year = {1998}
}

@article{stiller2007perspective,
title = {Perspective-taking and memory capacity predict social network size},
journal = {Social Networks},
volume = {29},
number = {1},
pages = {93-104},
year = {2007},
issn = {0378-8733},
doi = {https://doi.org/10.1016/j.socnet.2006.04.001},
url = {https://www.sciencedirect.com/science/article/pii/S0378873306000128},
author = {James Stiller and R.I.M. Dunbar}
}

@article{ogrady2015ease,
title = {The ease and extent of recursive mindreading, across implicit and explicit tasks},
journal = {Evolution and Human Behavior},
volume = {36},
number = {4},
pages = {313-322},
year = {2015},
issn = {1090-5138},
doi = {https://doi.org/10.1016/j.evolhumbehav.2015.01.004},
url = {https://www.sciencedirect.com/science/article/pii/S1090513815000148},
author = {Cathleen O’Grady and Christian Kliesch and Kenny Smith and Thomas C. Scott-Phillips}
}

@book{fagin1995reasoning,
  title     = {Reasoning About Knowledge},
  author    = {Fagin, Ronald and Halpern, Joseph Y. and Moses, Yoram and Vardi, Moshe Y.},
  publisher = {MIT Press},
  year      = {1995},
  doi       = {10.7551/mitpress/5803.001.0001}
}

@inproceedings{baltag1998logic,
author = {Baltag, Alexandru and Moss, Lawrence S. and Solecki, Slawomir},
title = {The logic of public announcements, common knowledge, and private suspicions},
year = {1998},
isbn = {1558605630},
publisher = {Morgan Kaufmann Publishers Inc.},
address = {San Francisco, CA, USA},
booktitle = {Proceedings of the 7th Conference on Theoretical Aspects of Rationality and Knowledge},
pages = {43–56},
numpages = {14},
location = {Evanston, Illinois},
series = {TARK '98}
}

@article{kosinski2024evaluating,
author = {Michal Kosinski },
title = {Evaluating large language models in theory of mind tasks},
journal = {Proceedings of the National Academy of Sciences},
volume = {121},
number = {45},
pages = {e2405460121},
year = {2024},
doi = {10.1073/pnas.2405460121},
URL = {https://www.pnas.org/doi/abs/10.1073/pnas.2405460121},
eprint = {https://www.pnas.org/doi/pdf/10.1073/pnas.2405460121}}

@inproceedings{le2019revisiting,
    title = "Revisiting the Evaluation of Theory of Mind through Question Answering",
    author = "Le, Matthew  and
      Boureau, Y-Lan  and
      Nickel, Maximilian",
    editor = "Inui, Kentaro  and
      Jiang, Jing  and
      Ng, Vincent  and
      Wan, Xiaojun",
    booktitle = "Proceedings of the 2019 Conference on Empirical Methods in Natural Language Processing and the 9th International Joint Conference on Natural Language Processing ({EMNLP}-{IJCNLP})",
    month = nov,
    year = "2019",
    address = "Hong Kong, China",
    publisher = "Association for Computational Linguistics",
    url = "https://aclanthology.org/D19-1598/",
    doi = "10.18653/v1/D19-1598",
    pages = "5872--5877"
}

@inproceedings{arodi2021textual,
    title = "Textual Time Travel: A Temporally Informed Approach to Theory of Mind",
    author = "Arodi, Akshatha  and
      Cheung, Jackie Chi Kit",
    editor = "Moens, Marie-Francine  and
      Huang, Xuanjing  and
      Specia, Lucia  and
      Yih, Scott Wen-tau",
    booktitle = "Findings of the {A}ssociation for {C}omputational {L}inguistics: {EMNLP} 2021",
    month = nov,
    year = "2021",
    address = "Punta Cana, Dominican Republic",
    publisher = "Association for Computational Linguistics",
    url = "https://aclanthology.org/2021.findings-emnlp.351/",
    doi = "10.18653/v1/2021.findings-emnlp.351",
    pages = "4162--4172"
}

@inproceedings{wu2023hitom,
    title = "Hi-{T}o{M}: A Benchmark for Evaluating Higher-Order Theory of Mind Reasoning in Large Language Models",
    author = "Wu, Yufan  and
      He, Yinghui  and
      Jia, Yilin  and
      Mihalcea, Rada  and
      Chen, Yulong  and
      Deng, Naihao",
    editor = "Bouamor, Houda  and
      Pino, Juan  and
      Bali, Kalika",
    booktitle = "Findings of the {A}ssociation for {C}omputational {L}inguistics: {EMNLP} 2023",
    month = dec,
    year = "2023",
    address = "Singapore",
    publisher = "Association for Computational Linguistics",
    url = "https://aclanthology.org/2023.findings-emnlp.717/",
    doi = "10.18653/v1/2023.findings-emnlp.717",
    pages = "10691--10706"
}

@inproceedings{kim2023fantom,
    title = "{FANT}o{M}: A Benchmark for Stress-testing Machine Theory of Mind in Interactions",
    author = "Kim, Hyunwoo  and
      Sclar, Melanie  and
      Zhou, Xuhui  and
      Bras, Ronan  and
      Kim, Gunhee  and
      Choi, Yejin  and
      Sap, Maarten",
    editor = "Bouamor, Houda  and
      Pino, Juan  and
      Bali, Kalika",
    booktitle = "Proceedings of the 2023 Conference on Empirical Methods in Natural Language Processing",
    month = dec,
    year = "2023",
    address = "Singapore",
    publisher = "Association for Computational Linguistics",
    url = "https://aclanthology.org/2023.emnlp-main.890/",
    doi = "10.18653/v1/2023.emnlp-main.890",
    pages = "14397--14413"
}

@inproceedings{xu2024opentom,
    title = "{O}pen{T}o{M}: A Comprehensive Benchmark for Evaluating Theory-of-Mind Reasoning Capabilities of Large Language Models",
    author = "Xu, Hainiu  and
      Zhao, Runcong  and
      Zhu, Lixing  and
      Du, Jinhua  and
      He, Yulan",
    editor = "Ku, Lun-Wei  and
      Martins, Andre  and
      Srikumar, Vivek",
    booktitle = "Proceedings of the 62nd Annual Meeting of the {A}ssociation for {C}omputational {L}inguistics (Volume 1: Long Papers)",
    month = aug,
    year = "2024",
    address = "Bangkok, Thailand",
    publisher = "Association for Computational Linguistics",
    url = "https://aclanthology.org/2024.acl-long.466/",
    doi = "10.18653/v1/2024.acl-long.466",
    pages = "8593--8623"
}

@inproceedings{chen2024tombench,
    title = "{T}o{MB}ench: Benchmarking Theory of Mind in Large Language Models",
    author = "Chen, Zhuang  and
      Wu, Jincenzi  and
      Zhou, Jinfeng  and
      Wen, Bosi  and
      Bi, Guanqun  and
      Jiang, Gongyao  and
      Cao, Yaru  and
      Hu, Mengting  and
      Lai, Yunghwei  and
      Xiong, Zexuan  and
      Huang, Minlie",
    editor = "Ku, Lun-Wei  and
      Martins, Andre  and
      Srikumar, Vivek",
    booktitle = "Proceedings of the 62nd Annual Meeting of the {A}ssociation for {C}omputational {L}inguistics (Volume 1: Long Papers)",
    month = aug,
    year = "2024",
    address = "Bangkok, Thailand",
    publisher = "Association for Computational Linguistics",
    url = "https://aclanthology.org/2024.acl-long.847/",
    doi = "10.18653/v1/2024.acl-long.847",
    pages = "15959--15983"
}

@ARTICLE{street2024llms,
    
AUTHOR={Street, Winnie  and Siy, John Oliver  and Keeling, Geoff  and Baranes, Adrien  and Barnett, Benjamin  and McKibben, Michael  and Kanyere, Tatenda  and Lentz, Alison  and Arcas, Blaise Agüera y  and Dunbar, Robin I. M. },
           
TITLE={LLMs achieve adult human performance on higher-order theory of mind tasks},
          
JOURNAL={Frontiers in Human Neuroscience},
          
VOLUME={Volume 19 - 2025},
  
YEAR={2026},
  
URL={https://www.frontiersin.org/journals/human-neuroscience/articles/10.3389/fnhum.2025.1633272},
  
DOI={10.3389/fnhum.2025.1633272},
  
ISSN={1662-5161}}

@inproceedings{sclar2025explore,
title={Explore Theory of Mind: program-guided adversarial data generation for theory of mind reasoning},
author={Melanie Sclar and Jane Dwivedi-Yu and Maryam Fazel-Zarandi and Yulia Tsvetkov and Yonatan Bisk and Yejin Choi and Asli Celikyilmaz},
booktitle={The Thirteenth International Conference on Learning Representations},
year={2025},
url={https://openreview.net/forum?id=246rHKUnnf}
}

@inproceedings{hou2024timetom,
    title = "{T}ime{T}o{M}: Temporal Space is the Key to Unlocking the Door of Large Language Models' Theory-of-Mind",
    author = "Hou, Guiyang  and
      Zhang, Wenqi  and
      Shen, Yongliang  and
      Wu, Linjuan  and
      Lu, Weiming",
    editor = "Ku, Lun-Wei  and
      Martins, Andre  and
      Srikumar, Vivek",
    booktitle = "Findings of the {A}ssociation for {C}omputational {L}inguistics: {ACL} 2024",
    month = aug,
    year = "2024",
    address = "Bangkok, Thailand",
    publisher = "Association for Computational Linguistics",
    url = "https://aclanthology.org/2024.findings-acl.685/",
    doi = "10.18653/v1/2024.findings-acl.685",
    pages = "11532--11547"
}

@misc{lei2026mind,
      title={Mind the Perspective: Let's Reason Recursively for Theory of Mind}, 
      author={Chao Lei and Guang Hu and Meng Yang and Yanbei Jiang and Nir Lipovetzky},
      year={2026},
      eprint={2606.11724},
      archivePrefix={arXiv},
      primaryClass={cs.AI},
      url={https://arxiv.org/abs/2606.11724}, 
}

@inproceedings{wilf2023think,
    title = "Think Twice: Perspective-Taking Improves Large Language Models' Theory-of-Mind Capabilities",
    author = "Wilf, Alex  and
      Lee, Sihyun  and
      Liang, Paul Pu  and
      Morency, Louis-Philippe",
    editor = "Ku, Lun-Wei  and
      Martins, Andre  and
      Srikumar, Vivek",
    booktitle = "Proceedings of the 62nd Annual Meeting of the {A}ssociation for {C}omputational {L}inguistics (Volume 1: Long Papers)",
    month = aug,
    year = "2024",
    address = "Bangkok, Thailand",
    publisher = "Association for Computational Linguistics",
    url = "https://aclanthology.org/2024.acl-long.451/",
    doi = "10.18653/v1/2024.acl-long.451",
    pages = "8292--8308"
}

@inproceedings{wei2022chain,
title={Chain of Thought Prompting Elicits Reasoning in Large Language Models},
author={Jason Wei and Xuezhi Wang and Dale Schuurmans and Maarten Bosma and brian ichter and Fei Xia and Ed H. Chi and Quoc V Le and Denny Zhou},
booktitle={Advances in Neural Information Processing Systems},
editor={Alice H. Oh and Alekh Agarwal and Danielle Belgrave and Kyunghyun Cho},
year={2022},
url={https://openreview.net/forum?id=_VjQlMeSB_J}
}

@inproceedings{prakash2025lookbacks,
title={Language Models Use Lookbacks to Track Beliefs},
author={Nikhil Prakash and Natalie Shapira and Arnab Sen Sharma and Christoph Riedl and Yonatan Belinkov and Tamar Rott Shaham and David Bau and Atticus Geiger},
booktitle={The Fourteenth International Conference on Learning Representations},
year={2026},
url={https://openreview.net/forum?id=6gO6KTRMpG}
}

@misc{openai2026gpt6,
  title        = {{GPT-6} system card},
  author       = {{OpenAI}},
  year         = {2026},
  howpublished = {Model card},
  url          = {https://deploymentsafety.openai.com/gpt-6-astra},
  note         = {Accessed October 2026}
}

@misc{anthropic2026claude,
  title        = {System card: {Claude Opus 5.5}},
  author       = {{Anthropic}},
  year         = {2026},
  month        = sep,
  howpublished = {System card},
  url          = {https://anthropic.com/claude-opus-5-5-system-card},
  note         = {Accessed October 2026}
}

@misc{google2026gemini3,
  title        = {{Gemini 3 flash} model card},
  author       = {{Google DeepMind}},
  year         = {2026},
  howpublished = {Model card},
  url          = {https://storage.googleapis.com/deepmind-media/Model-Cards/Gemini-3-Flash-Model-Card.pdf},
  note         = {Gemini 3 Flash; accessed October 2026}
}

@misc{deepseek2026v4,
      title={DeepSeek-V4: Towards Highly Efficient Million-Token Context Intelligence}, 
      author={DeepSeek-AI and Anyi Xu and Bangcai Lin and Bing Xue and Bingxuan Wang and Bingzheng Xu and Bochao Wu and Bowei Zhang and Chaofan Lin and Chen Dong and Chenchen Ling and Chengda Lu and Chenggang Zhao and Chengqi Deng and Chengyu Hou and Chenhao Xu and Chenze Shao and Chong Ruan and Conner Sun and Damai Dai and Daya Guo and Dejian Yang and Deli Chen and Donghao Li and Dongjie Ji and Erhang Li and Fang Wei and Fangyun Lin and Fangzhou Yuan and Feiyu Xia and Fucong Dai and Guangbo Hao and Guanting Chen and Guoai Cao and Guolai Meng and Guowei Li and Han Yu and Han Zhang and Hanwei Xu and Hao Li and Haofen Liang and Haoling Zhang and Haoming Luo and Haoran Wei and Haotian Yuan and Haowei Zhang and Haowen Luo and Haoyu Chen and Haozhe Ji and Hengqing Zhang and Honghui Ding and Hongxuan Tang and Huanqi Cao and Huazuo Gao and Hui Qu and Hui Zeng and J Yang and JQ Zhu and Jia Luo and Jia Song and Jia Yu and Jialiang Huang and Jialu Cai and Jian Liang and Jiangting Zhou and Jiasheng Ye and Jiashi Li and Jiaxin Xu and Jiewen Hu and Jieyu Yang and Jin Chen and Jin Yan and Jingchang Chen and Jingli Zhou and Jingting Xiang and Jingyang Yuan and Jingyuan Cheng and Jingzi Zhou and Jinhua Zhu and Jiping Yu and Joseph Sun and Jun Ran and Junguang Jiang and Junjie Qiu and Junlong Li and Junmin Zheng and Junxiao Song and Kai Dong and Kaige Gao and Kang Guan and Kexing Zhou and Kezhao Huang and Kuai Yu and Lean Wang and Lecong Zhang and Lei Wang and Leyi Xia and Li Zhang and Liang Zhao and Lihua Guo and Lingxiao Luo and Linwang Ma and Linyan Zhu and Litong Wang and Liyu Cai and Liyue Zhang and Longhao Chen and MS Di and MY Xu and Max Mei and Miaojun Wang and Mingchuan Zhang and Minghua Zhang and Minghui Tang and Mingming Li and Mingxu Zhou and Minmin Han and Ning Wang and Panpan Huang and Panpan Wang and Peixin Cong and Peiyi Wang and Peng Zhang and Qiancheng Wang and Qihao Zhu and Qingyang Li and Qinyu Chen and Qiushi Du and Qiwei Jiang and Rui Tian and Ruifan Xu and Ruijie Lu and Ruiling Xu and Ruiqi Ge and Ruisong Zhang and Ruizhe Pan and Runji Wang and Runqian Chen and Runqiu Yin and Runxin Xu and Ruomeng Shen and Ruoyu Zhang and Ruyi Chen and SH Liu and Shanghao Lu and Shangmian Sun and Shangyan Zhou and Shanhuang Chen and Shaofei Cai and Shaoheng Nie and Shaoqing Wu and Shaoyuan Chen and Shengding Hu and Shengyu Liu and Shiqiang Hu and Shirong Ma and Shiyu Wang and Shuiping Yu and Shunfeng Zhou and Shuting Pan and Shuying Yu and Songyang Zhou and Tao Ni and Tao Yun and Tian Jin and Tian Pei and Tian Ye and Tianle Lin and Tianran Ji and Tianyi Cui and Tianyuan Yue and Tingting Yu and Tun Wang and W Zhang and WL Xiao and Wangding Zeng and Wei An and Weilin Zhao and Wen Liu and Wenfeng Liang and Wenjie Pang and Wenjing Luo and Wenjing Yao and Wenjun Gao and Wenkai Yang and Wenlve Huang and Wenqing Hou and Wentao Zhang and Wenting Ma and Xi Gao and Xiang He and Xiangwen Wang and Xianzu Wang and Xiao Bi and Xiaodong Liu and Xiaohan Wang and Xiaokang Chen and Xiaokang Zhang and Xiaotao Nie and Xiaowen Sun and Xiaoxiang Wang and Xin Cheng and Xin Liu and Xin Xie and Xingchao Liu and Xingchen Liu and Xingkai Yu and Xingyou Li and Xinyu Yang and Xinyu Zhang and Xu Chen and Xuanyu Wang and Xuecheng Su and Xueyin Chen and Xuheng Lin and Xuwei Fu and YC Yan and YQ Wang and YW Ma and Yanfeng Luo and Yang Zhang and Yanhong Xu and Yanru Ma and Yanwen Huang and Yao Li and Yao Li and Yao Xu and Yao Zhao and Yaofeng Sun and Yaohui Wang and Yi Qian and Yi Shao and Yi Yu and Yichao Zhang and Yifan Ding and Yifan Shi and Yijia Wu and Yiliang Xiong and Yiling Ma and Ying He and Ying Tang and Ying Zhou and Yingjia Luo and Yinmin Zhong and Yishi Piao and Yisong Wang and Yixiang Zhang and Yixiao Chen and Yixuan Tan and Yixuan Wei and Yiyang Ma and Yiyuan Liu and Yonglun Yang and Yongqiang Guo and Yongtong Wu and Yu Wu and YuKun Li and Yuan Cheng and Yuan Ou and Yuanfan Xu and Yuanhao Li and Yuduan Wang and Yuehan Yang and Yuer Xu and Yuhan Wu and Yuhao Meng and Yuheng Zou and Yukun Zha and Yunfan Xiong and Yupeng Chen and Yuping Lin and Yuqian Cao and Yuqian Wang and Yushun Zhang and Yuting Yan and Yutong Lin and Yuxian Gu and Yuxiang Luo and Yuxiang You and Yuxuan Liu and Yuxuan Zhou and Yuyang Zhou and Yuzhen Huang and ZF Wu and Zehao Wang and Zehua Zhao and Zehui Ren and Zekai Zhang and Zhangli Sha and Zhe Fu and Zhe Ju and Zhean Xu and Zhenda Xie and Zhengyan Zhang and Zheren Gao and Zhewen Hao and Zhibin Gou and Zhicheng Ma and Zhigang Yan and Zhihong Shao and Zhixian Huang and Zhixuan Chen and Zhiyu Wu and Zhizhou Ren and Zhongyu Wu and Zhuoshu Li and Zhuping Zhang and Zian Xu and Zihao Wang and Zihua Qu and Zihui Gu and Zijia Zhu and Zilin Li and Zipeng Zhang and Ziwei Xie and Ziyi Gao and Ziyi Wan and Zizheng Pan and Zongqing Yao},
      year={2026},
      eprint={2606.19348},
      archivePrefix={arXiv},
      primaryClass={cs.CL},
      url={https://arxiv.org/abs/2606.19348}, 
}

@misc{yang2025qwen3,
      title={Qwen3 Technical Report}, 
      author={An Yang and Anfeng Li and Baosong Yang and Beichen Zhang and Binyuan Hui and Bo Zheng and Bowen Yu and Chang Gao and Chengen Huang and Chenxu Lv and Chujie Zheng and Dayiheng Liu and Fan Zhou and Fei Huang and Feng Hu and Hao Ge and Haoran Wei and Huan Lin and Jialong Tang and Jian Yang and Jianhong Tu and Jianwei Zhang and Jianxin Yang and Jiaxi Yang and Jing Zhou and Jingren Zhou and Junyang Lin and Kai Dang and Keqin Bao and Kexin Yang and Le Yu and Lianghao Deng and Mei Li and Mingfeng Xue and Mingze Li and Pei Zhang and Peng Wang and Qin Zhu and Rui Men and Ruize Gao and Shixuan Liu and Shuang Luo and Tianhao Li and Tianyi Tang and Wenbiao Yin and Xingzhang Ren and Xinyu Wang and Xinyu Zhang and Xuancheng Ren and Yang Fan and Yang Su and Yichang Zhang and Yinger Zhang and Yu Wan and Yuqiong Liu and Zekun Wang and Zeyu Cui and Zhenru Zhang and Zhipeng Zhou and Zihan Qiu},
      year={2025},
      eprint={2505.09388},
      archivePrefix={arXiv},
      primaryClass={cs.CL},
      url={https://arxiv.org/abs/2505.09388}, 
}

@misc{google2026gemma4,
      title={Gemma 4 Technical Report}, 
      author={Gemma Team and Sherif El Abd and Vaibhav Aggarwal and Robin Algayres and Alek Andreev and Olivier Bachem and Ian Ballantyne and Cormac Brick and Victor Cărbune and Michelle Casbon and Mayank Chaturvedi and Aditya Chawla and Victor Cotruta and Alice Coucke and Phil Culliton and Robert Dadashi and Lucas Dixon and Mohamed Elhawaty and Utku Evci and Clément Farabet and Johan Ferret and Filippo Galgani and Sertan Girgin and Jean-Bastien Grill and Maarten Grootendorst and Jiaxian Guo and Cassidy Hardin and Yanzhang He and Steven M. Hernandez and Omri Homburger and Léonard Hussenot and Juyeong Ji and Armand Joulin and Aishwarya Kamath and Parnian Kassraie and Olivier Lacombe and Preethi Lahoti and Gaël Liu and Gus Martins and Luciano Martins and Tatiana Matejovicova and Ramona Merhej and Nikola Momchev and Sneha Mondal and Ryan Mullins and Sindhu Raghuram Panyam and Shreya Pathak and Sarah Perrin and André Susano Pinto and Etienne Pot and Angéline Pouget and Alexandre Ramé and Sabela Ramos and Douglas Reid and David Rim and Morgane Rivière and Karsten Roth and Louis Rouillard and Omar Sanseviero and Pier Giuseppe Sessa and Shane Settle and Danila Sinopalnikov and Sara Smoot and Piotr Stanczyk and Andreas Steiner and Lawrence Stewart and Ilya Tolstikhin and Michael Tschannen and Anton Tsitsulin and Nino Vieillard and Renjie Wu and Pingmei Xu and Haichuan Yang and Edouard Yvinec and Biao Zhang and Li Zhang and Joe Zou and Nicolas Aagnes and Abdelrahman Abdelhamed and Jakub Adamek and Shivani Agrawal and Shubham Agrawal and Ibrahim Alabdulmohsin and Jean Baptiste Alayrac and Uri Alon and Chandramouli Amarnath and Ankesh Anand and Chrysovalantis Anastasiou and Setareh Ariafar and François-Xavier Aubet and Kyriakos Axiotis and Federico Barbero and Joelle Barral and Alexei Bendebury and Urs Bergmann and Stanley Bileschi and Kat Black and Mathieu Blondel and Sebastian Borgeaud and Arthur Bražinskas and Ryan Burnell and Robert Busa-Fekete and Mu Cai and Daniele Calandriello and Glenn Cameron and Charlotte Caucheteux and Rahma Chaabouni and Garima Chadha and Jetha Chan and Blake Jianhang Chen and Jesse Chen and Lin Chen and Xu Chen and Derek Cheng and Tzu-hsiang Chien and Nikolai Chinaev and Yi Chou and Zhaohui Chu and Benjamin Coleman and Pooja Consul and Sam Conway-Rahman and Scott Crowell and Dylan Cutler and Vivek Dani and Samira Daruki and Anil Das and Daniel Deutsch and Nishanth Dikkala and Li Ding and Qiuhan Ding and Shenil Dodhia and Konstantin Donhauser and Tulsee Doshi and Anca Dragan and Alex Druinsky and Sahil Dua and Zoltan Egyed and Danielle Eisenbud and Daniel Eppens and Cindy Fan and Bahare Fatemi and Yassir Fathullah and Vlad Feinberg and Milen Ferev and Sebastian Flennerhag and Takumi Fujimoto and João Gabriel Oliveira and Isaac Galatzer-Levy and João Gante and Simon Geisler and Soham Ghosal and Antonious M. Girgis and Tamara von Glehn and Alec Go and Alhaad Gokhale and Alex Grills and Yiming Gu and Mayank Gupta and Pramod Gupta and Guru Guruganesh and Raia Hadsell and Hamza Harkous and Jitendra Harlalka and Demis Hassabis and Anja Hauth and Joe Heyward and Arian Hosseini and Chih-Yang Hsia and I-Hung Hsu and Xiaopeng Huang and Yangsibo Huang and Kevin Hui and Adrian Hutter and Te I and Fotis Iliopoulos and Advait Jain and Ganesh Jawahar and Ziwei Ji and Qilin Jin and Melvin Johnson and Kandarp Joshi and Arun Kandoor and Wang-Cheng Kang and Koray Kavukcuoglu and Mehran Kazemi and Kathleen Kenealy and Amr Khalifa and Phoebe Kirk and Ivan Korotkov and Suraj Kothawade and Vitaly Kovalev and Neel Kovelamudi and Adam Kraft and Ravin Kumar and Vivek Kumar and Harish Kuppam and Justin Lannin and Chen-Yu Lee and Seungji Lee and Dmitry Lepikhin and Alon Levkovitch and Dongdong Li and Qiujia Li and Valentin Liévin and Ethan Lin and Ziqian Lin and Casper Liu and Tianlin Liu and Tianqi Liu and Xin Liu and Ivan Lobov and Mayank Lunayach and Min Ma and Gagan Madan and Andrii Maksai and Eric Malmi and Michal Matuszak and Daniel McDuff and Gaurav Menghani and Maciej Mikuła and Daniil Mirylenka and Karolis Misiunas and Vedant Misra and Andreea Mitran and Kareem Mohamed and Maksim Mukha and Eric Noland and James O'Donnell and Brendan O'Donoghue and Kate Olszewska and Bernett Orlando and Wanqiong Pan and Rina Panigrahy and Unnati Parekh and Nicolas Perez-Nieves and Chunjong Park and Eric Paskie and Liqian Peng and Bryce Petrini and Slav Petrov and Jonas Pfeiffer and Bilal Piot and Martyna Plomecka and Siim Poder and Octavio Ponce and Arijit Pramanik and David Racz and Anish Rajan and Michelle Ramanovich and Anand Rao and Marvin Ritter and Vitor Rodrigues and Evan Rosen and Mikołaj Rybiński and Noveen Sachdeva and Michaël E. Sander and Rohit Sathyanarayana and Sagar Savla and Samuel Schmidgall and Tal Schuster and George Scrivener and Benoit Seguin and Andrew Sellergren and Aliaksei Severyn and Izhak Shafran and Dhruv Shah and Bobak Shahriari and Yuan Shangguan and Ashish Shenoy and Pradeep Shenoy and Rakesh Shivanna and Pauline Sho and Lucas Spangher and Wojciech Stokowiec and Tim Strother and Yao Su and Yinghao Sun and Mukund Sundararajan and Andrea Tacchetti and Mor Hazan Taege and Pouya Tafti and Jean Tarbouriech and Chetan Tekur and Shantanu Thakoor and Rahul Thapa and Madeleine Traverse and Lenart Treven and Tao Tu and Chien Te Tung and Çağlar Ünlü and Petar Veličković and Malini Pooni Venkat and Sagar Gubbi Venkatesh and Vidya Venkiteswaran and Francesco Visin and Alex Vitvitskyi and Kiran Vodrahalli and Weiyi Wang and Xin Wang and Tris Warkentin and Jan Wassenberg and John Wieting and Cindy Wu and Lechao Xiao and Hao Xu and Yuhui Xu and Fuzhao Xue and Arun Yadav and Jun Yan and Antoine Yang and Lin Yang and Ming-Hsuan Yang and Ziyu Ying and Jae Hyeon Yoo and Morteza Zadimoghaddam and Sajjad Zafar and Fred Zhang and Jiageng Zhang and Jianyi Zhang and Xiaofan Zhang and Chao Zhao and David Zhou and Chen Zou},
      year={2026},
      eprint={2607.02770},
      archivePrefix={arXiv},
      primaryClass={cs.CL},
      url={https://arxiv.org/abs/2607.02770}, 
}

@article{wilson2023recursive,
  title={Is recursive “mindreading” really an exception to limitations on recursive thinking?},
  author={Wilson, Ross and Hruby, Ales and Perez-Zapata, Daniel and van der Kleij, Sanne W and Apperly, Ian A},
  journal={Journal of Experimental Psychology: General},
  volume={152},
  number={5},
  pages={1454},
  year={2023},
  publisher={American Psychological Association},
  doi     = {10.1037/xge0001322}
}

@inproceedings{balepur2024artifacts,
    title = "Artifacts or Abduction: How Do {LLM}s Answer Multiple-Choice Questions Without the Question?",
    author = "Balepur, Nishant  and
      Ravichander, Abhilasha  and
      Rudinger, Rachel",
    editor = "Ku, Lun-Wei  and
      Martins, Andre  and
      Srikumar, Vivek",
    booktitle = "Proceedings of the 62nd Annual Meeting of the {A}ssociation for {C}omputational {L}inguistics (Volume 1: Long Papers)",
    month = aug,
    year = "2024",
    address = "Bangkok, Thailand",
    publisher = "Association for Computational Linguistics",
    url = "https://aclanthology.org/2024.acl-long.555/",
    doi = "10.18653/v1/2024.acl-long.555",
    pages = "10308--10330"
}

@inproceedings{zheng2024robust,
title={Large Language Models Are Not Robust Multiple Choice Selectors},
author={Chujie Zheng and Hao Zhou and Fandong Meng and Jie Zhou and Minlie Huang},
booktitle={The Twelfth International Conference on Learning Representations},
year={2024},
url={https://openreview.net/forum?id=shr9PXz7T0}
}

@inproceedings{meng2022locating,
title={Locating and Editing Factual Associations in {GPT}},
author={Kevin Meng and David Bau and Alex J Andonian and Yonatan Belinkov},
booktitle={Advances in Neural Information Processing Systems},
editor={Alice H. Oh and Alekh Agarwal and Danielle Belgrave and Kyunghyun Cho},
year={2022},
url={https://openreview.net/forum?id=-h6WAS6eE4}
}

@inproceedings{geva2023dissecting,
    title = "Dissecting Recall of Factual Associations in Auto-Regressive Language Models",
    author = "Geva, Mor  and
      Bastings, Jasmijn  and
      Filippova, Katja  and
      Globerson, Amir",
    editor = "Bouamor, Houda  and
      Pino, Juan  and
      Bali, Kalika",
    booktitle = "Proceedings of the 2023 Conference on Empirical Methods in Natural Language Processing",
    month = dec,
    year = "2023",
    address = "Singapore",
    publisher = "Association for Computational Linguistics",
    url = "https://aclanthology.org/2023.emnlp-main.751/",
    doi = "10.18653/v1/2023.emnlp-main.751",
    pages = "12216--12235"
}

\clearpage
\appendix
\begin{table*}[b]
\centering
\small
\setlength{\tabcolsep}{4pt}
\resizebox{0.99\textwidth}{!}{
\begin{tabular}{@{}l l c l l l@{}}
\toprule
\textbf{Benchmark} & \textbf{Source} & \textbf{Max level} & \textbf{Size} & \textbf{Format} & \textbf{Order convention} \\
\midrule
ToMi$^\dagger$ & \citet{le2019revisiting} & 2 & 5,994 test questions & open-ended & nesting depth \\
Hi-ToM$^\dagger$ & \citet{wu2023hitom} & 4 & 120 stories, 600 unique questions & 15-way choice & nesting depth \\
FANToM & \citet{kim2023fantom} & 2 & 256 conversations, 870 short sets & choice, free-form, list & nesting depth \\
OpenToM & \citet{xu2024opentom} & 2 & 696 narratives, 16,008 questions & 2/3-way choice & nesting depth \\
ToMBench & \citet{chen2024tombench} & 2 & 2,860 items & 4-way choice & task taxonomy \\
ExploreToM$^\dagger$ & \citet{sclar2025explore} & 2 & 619 stories, 13,309 questions (sample) & open-ended, yes/no & nesting depth \\
MoToMQA$^\dagger$ & \citet{street2024llms} & 6 & 7 stories, 70 ToM statements & true/false & respondent included in level \\
\midrule
\bench & this work & 5 & 200 realizations, 1,800 questions & open-ended, optional choice & certified effective order \\
\bottomrule
\end{tabular}
}
\caption{Benchmarks with nested-belief questions. Maximum levels follow each source's convention. $^\dagger$Audited in this work; Appendix~\ref{app:audit} describes the audited subsets.}
\label{tab:landscape}
\end{table*}

\section{Benchmark Audit}
\label{app:audit}

Table~\ref{tab:landscape} lists the benchmarks with nested-belief questions. For each audited subset, we encode the stories into perspective records and apply the support rule of Sec.~\ref{sec:formulation} to every belief prefix. As an encoding check, we compare the solver's answers with the released labels.

\paragraph{ToMi.}
We encode the test split of the balanced release programmatically and audit its 1,998 second-order questions. Mean effective order is 1.36, and 36.2\% of the questions have full effective order. The other 1,275 questions have collapsed layers: 766 through idle gaps and 509 through subsumed perspectives. Our solver matches the released gold answer on all 550 ``ToM'' questions and on 68.0\% of the 1,448 ``no-ToM'' questions. This gap is consistent with the label inconsistencies reported by \citet{arodi2021textual}.

\paragraph{ExploreToM.}
The released sample has 619 story structures and 13,309 questions: 3,278 first-order, 9,184 second-order, and 847 memory questions. We compile its action scripts into perspective records and audit the 3,905 location questions (607 first-order and 3,298 second-order). On this subset, the solver matches the released answers on 98.5\% of questions. For the second-order location questions, mean effective order is 1.07 and 6.6\% have full effective order. Of the remaining 3,080 questions, 2,881 contain idle gaps and 199 contain subsumed perspectives.

\paragraph{Hi-ToM.}
Hi-ToM has 120 unique questions at each published level from zero to four. Every question appears twice, once in a chain-of-thought prompt (CoTP) and once in a vanilla prompt (VP). For 138 questions the two copies carry different gold answers (0, 0, 35, 48, and 55 at levels zero to four), which leaves 120, 120, 85, 72, and 65 questions with consistent labels. Answer-based scoring uses the released labels. The effective-order audit does not depend on gold answers, so it covers all 600 unique questions. Table~\ref{tab:hitom_audit} reports it for orders one to four.

\paragraph{MoToMQA.}
MoToMQA consists of hand-written stories without action logs, and its statements involve desires, intentions, and emotions besides beliefs. We therefore report its audit here rather than in Table~\ref{tab:audit}. MoToMQA's published levels two to six count the respondent and every mental-state verb, including those in coordinated clauses, so its 70 ToM statements have syntactic orders 1, 2, 3, 3--4, and 3--4. We annotate the event that supports each belief prefix with an LLM-based annotator. Mean effective order at the five published levels is 1.00, 1.07, 1.50, 1.50, and 2.00. No statement above published level three has full effective order, and the highest effective order observed is three. In 10 of the 21 true statements with $k\geq3$, the story states the nested mental state explicitly, so a single sentence settles several prefixes at once.

\FloatBarrier

\section{\bench Construction}
\label{app:construction}
Each prototype combines one or more of the 10 scenario types in Table~\ref{tab:scenarios}.

Each narrative realization has one control at each syntactic order from two to five, giving 400 controls across the 100 narrative realizations. By collapse type, 120 contain subsumed perspectives, 130 idle gaps, and 100 both; the remaining 50 have full effective order but an answer that coincides with a lower-order alternative. Their mean effective orders at orders two to five are 1.20, 1.80, 2.70, and 3.10. 

Each of the 100 underlying interactions is realized in two story formats: a prose narrative and a script-style dialogue with character lines and stage directions. The two realizations of an interaction share the same questions, gold answers, belief chains, and labels, and differ only in how the story is presented. Every question is provided with open-ended answers and as a seven-option multiple-choice item with the same stem and gold answer (Appendix~\ref{app:mc}). \bench is bilingual: every story and question is provided in Chinese and English with identical belief chains, orders, labels, and option positions, giving 400 stories and 3,600 questions across the two languages.

\begin{table*}[t]
\centering
\small
\setlength{\tabcolsep}{5pt}
\renewcommand{\arraystretch}{1.15}
\resizebox{0.99\textwidth}{!}{
\begin{tabular}{@{}p{3.4cm} l p{9.6cm}@{}}
\toprule
\textbf{Family} & \textbf{Type} & \textbf{Definition} \\
\midrule
\multirow{2}{=}{Asymmetric Epistemic Access}
 & present but inattentive & An agent is physically present during an event but does not attend to it and therefore does not acquire the relevant information. \\
\cmidrule(l){2-3}
 & secret observation & An agent observes an event without one or more other agents knowing that the observation occurred. \\
\midrule
\multirow{2}{=}{Temporal Presence Dynamics}
 & leave and return & An agent leaves before a relevant event and later returns without direct access to what occurred during the absence. \\
\cmidrule(l){2-3}
 & late joining & An agent enters after earlier events have occurred and lacks the information available before arrival. \\
\midrule
Strategic Information Presentation
 & deceptive communication & An agent intentionally provides false or misleading information to influence another agent's belief. \\
\midrule
\multirow{2}{=}{Unintentional Epistemic Error}
 & observation misperception & An agent observes an event but incorrectly interprets what happened. \\
\cmidrule(l){2-3}
 & unintentional misreport & An agent communicates incorrect information because of an error rather than an intention to deceive. \\
\midrule
\multirow{2}{=}{Asynchronous Epistemic Updating}
 & delayed meta-information & An agent learns only later that another agent previously observed, knew, or communicated a relevant fact. \\
\cmidrule(l){2-3}
 & nonmonotonic trace & Later evidence causes an agent to revise, retract, or reinstate an earlier belief. \\
\midrule
Cyclic Query
 & cyclic query & A query follows a nested belief chain that returns to an earlier agent, such as ``A believes that B believes that A believes something.'' \\
\bottomrule
\end{tabular}
}
\caption{The ten scenario types of \bench, grouped into six families. A prototype can combine several types.}
\label{tab:scenarios}
\end{table*}

\FloatBarrier


\section{Human Study}
\label{app:human}

\paragraph{Procedure.}
Participants were recruited through an Chinese online survey panel in September 2026 and completed a web questionnaire with three stories, drawn from prototypes P1 (``the painting in the garden''), P3 (``the museum tickets''), and P9 (``the passport and the power bank''). Each story was followed by its five certified questions (orders one to five) and then its four controls (orders two to five). For each question, participants typed an answer and rated their confidence on a five-point scale. 

\paragraph{Participants.}
Twenty participants answered all 27 questions: 11 female and 9 male; 12 aged 21--30, 6 aged 31--40, and 2 aged 41--50; 17 with a bachelor's degree, 2 with a master's degree, and 1 with a college diploma. Mean completion time was 24.0 minutes. 

\paragraph{Accuracy.}
Table~\ref{tab:human} reports accuracy by syntactic order. On certified questions, accuracy is 58.3\%, 31.7\%, and 58.3\% at orders one to three, and falls to 6.7\% and 11.7\% at orders four and five. Controls are answered more accurately than certified questions at orders two and four, with the largest gap at order two (76.7\% versus 31.7\%).

\paragraph{Confidence.}
Mean confidence was 3.82 out of five. It was higher for correct than for incorrect answers (4.10 versus 3.61; paired permutation test, $p$ < 0.001). Accuracy was 24.7\%, 30.0\%, and 45.4\% at low, medium, and high confidence. Across participants, mean confidence did not correlate significantly with accuracy (Spearman $\rho$ = -0.14, $p$ = 0.55).

\paragraph{Ethics.}
The study was approved by the institutional review board of the authors' institution. Before starting, participants read a description of the study and gave informed consent; they could withdraw at any time. Each participant received monetary compensation for completing the questionnaire. We collected no identifying information beyond the demographic items above.

\begin{table}[t]
\centering
\small
\setlength{\tabcolsep}{4pt}
\resizebox{0.45\textwidth}{!}{
\begin{tabular}{@{}l ccccc c@{}}
\toprule
 & \textbf{1st} & \textbf{2nd} & \textbf{3rd} & \textbf{4th} & \textbf{5th} & \textbf{All} \\
\midrule
Certified & 58.3 & 31.7 & 58.3 & 6.7  & 11.7 & 33.3 \\
Control   & --   & 76.7 & 55.0 & 41.7 & 11.7 & 46.3 \\
\midrule
Overall   & 58.3 & 54.2 & 56.7 & 24.2 & 11.7 & 39.1 \\
\bottomrule
\end{tabular}
}
\caption{Human accuracy (\%) by syntactic order. }
\vspace{-0.2cm}
\label{tab:human}
\end{table}

\FloatBarrier

\section{Experimental Details}
\label{app:impl}

We queried the closed models in September 2026, using the identifiers gpt-6-sol, claude-opus-5-5, Gemini 3 Flash, and deepseek-v4-pro. The main runs do not request extended thinking. In the thinking runs, Claude Opus 5.5 uses adaptive thinking at high effort and Gemini 3 Flash uses its default alias. We ran Gemma 4 31B and Qwen3-32B with Hugging Face Transformers in bf16 on four NVIDIA A800 (80\,GB) GPUs, with greedy decoding and thinking disabled.

\FloatBarrier

\section{Additional Results}
\label{app:results}

\subsection{Full-Benchmark Results}
\label{app:full}

Across all 1,800 questions (certified questions and controls, narrative and dialogue realizations), CoT reaches 59.4\%, 59.1\%, 57.7\%, and 45.6\% with GPT-6 Sol, Claude Opus 5.5, Gemini 3 Flash, and DeepSeek-V4-Pro. SimToM~\citep{wilf2023think} reaches 57.6\%, 59.2\%, 40.8\%, and 40.7\%, and \ours reaches 71.2\%, 70.4\%, 72.9\%, and 72.3\%. On the controls of the 100 narrative realizations, \ours reaches 67.8\%, 68.2\%, 68.2\%, and 67.0\% in the same model order. Certified results are provided in Table~\ref{tab:main}.

\subsection{Inference Time}
\label{app:time}

Per narrative realization, \ours takes 217, 98, 4, and 29 seconds with GPT-6 Sol, Claude Opus 5.5, Gemini 3 Flash, and DeepSeek-V4-Pro. TimeToM takes 808, 278, 442, and 1,067 seconds, and RecToM takes 827, 268, 455, and 1,258 seconds.

\subsection{Open-Ended and Multiple-Choice Answers}
\label{app:mc}

\paragraph{Format.}
Each multiple-choice item reuses the story and question of an open-ended item and lists seven options: the gold answer, five distractors, and ``I don't know''. The distractors come from a per-prototype pool of states that the story mentions for the same target variable, so they can include the target's actual state or the answer to a shorter belief prefix. For each of the nine question positions, both the gold answer and ``I don't know'' are spread evenly across the seven labels over the 100 interactions, and ``I don't know'' never occupies the gold position. Models answer the 900 questions of the 100 narrative realizations. Appendix~\ref{app:mech-mc} analyzes how open-weight models answer this format.

\begin{table}[t]
\centering
\small
\setlength{\tabcolsep}{3pt}
\resizebox{0.49\textwidth}{!}{
\begin{tabular}{@{}l cc c cc@{}}
\toprule
\textbf{Model} & \textbf{Open} & \textbf{MC} & $p$ & \textbf{Certified} & \textbf{Controls} \\
\midrule
Claude Opus 5.5 & 34.0 & 92.4 & $<10^{-100}$ & 35.2 $\to$ 89.2 & 32.5 $\to$ 96.5 \\
GPT-6 Sol       & 55.4 & 58.3 & 0.12          & 55.4 $\to$ 55.6 & 55.5 $\to$ 61.8 \\
Gemini 3 Flash  & 34.2 & 44.0 & $<10^{-10}$  & 34.0 $\to$ 45.0 & 34.5 $\to$ 42.8 \\
DeepSeek-V4-Pro & 36.2 & 30.0 & $2\times10^{-4}$ & 34.2 $\to$ 33.2 & 38.8 $\to$ 26.0 \\
\bottomrule
\end{tabular}
}
\caption{Open-ended and seven-option multiple-choice accuracy (\%) on the 900 questions of the 100 narrative realizations. \textbf{Open} and \textbf{MC}: accuracy in each format. $p$: paired McNemar test between the two formats. \textbf{Certified} and \textbf{Controls}: accuracy on each question subset, open-ended $\to$ multiple choice.}
\label{tab:mc}
\end{table}

\paragraph{Results.}
The effect of answer options depends on the model (Table~\ref{tab:mc}). Claude Opus 5.5 gains 58.4 points and reaches 96.5\% on controls. Gemini 3 Flash gains 9.8 points. DeepSeek-V4-Pro loses 6.2 points, mostly on controls. GPT-6 Sol gains 2.9 points, which is not significant under the paired test. Options can supply cues and allow elimination, and they can also introduce response biases. We therefore report open-ended accuracy as the primary measure of answering nested-belief queries without candidate answers.

\paragraph{Mechanistic analysis.}
Appendix~\ref{app:mech-mc} examines how options help the two open-weight models. Options raise accuracy from 33\% to 45\% for Gemma 4 31B and from 25\% to 37\% for Qwen3-32B, and accuracy on shortcut-resistant certified questions rises from 10\% to 24--29\%. Among correct multiple-choice answers, the paragraph that decides the answer is still the most influential one in 74--75\% of cases. The dependence on this paragraph weakens with order, however: hiding it costs Qwen3 only about 1 nat at fifth order. Correct answers to deep questions therefore draw more on the options, consistent with answering by elimination.

\FloatBarrier

\section{Mechanistic Analysis}
\label{app:mech}

\subsection{Experimental Setup}
\label{app:mech-setup}

We study Gemma 4 31B and Qwen3-32B in bf16 with greedy decoding on single questions. The analyses rely on two interventions. \emph{Segment hiding} blocks attention from the question and answer tokens to one story paragraph or prompt block, and measures the drop in the log-probability of the model's original answer in nats~\citep{geva2023dissecting}. We call a drop above 1 nat \emph{critical}. \emph{Activation patching} replaces the residual vector at a given layer and token position with the vector from another run and records how the answer changes~\citep{meng2022locating}. For a query $q$, the \emph{deciding record} is $E_{s_k}$, the record supporting the full prefix $p_k$. A question is \emph{shortcut-solvable} if its gold answer equals the answer to a shorter prefix, the outermost or innermost agent's own belief, or the target's actual state, and \emph{shortcut-resistant} otherwise.

\subsection{Direct-Answering Behavior}
\label{app:mech-behavior}

\begin{figure}[t]
\centering
\includegraphics[width=\columnwidth]{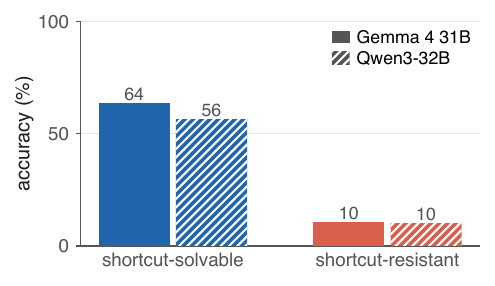}
\vspace{-0.5cm}
\caption{Direct-answering accuracy of Gemma 4 31B (solid) and Qwen3-32B (hatched) on certified questions of order $k\geq2$, split by whether the question is shortcut-solvable (Appendix~\ref{app:mech-setup}). Accuracy by order is in Fig.~\ref{fig:truepseudo}b, and wrong answers are broken down in Fig.~\ref{fig:mech-diag}a.}
\label{fig:mechA1}
\vspace{-0.2cm}
\end{figure}

On all 1,800 questions, Gemma and Qwen3 reach 33.2\% and 25.3\%. On certified questions of order $k\geq2$ that are shortcut-resistant, both reach 10\%, compared with 64\% and 56\% on shortcut-solvable ones (Fig.~\ref{fig:mechA1}). Among their errors on certified fourth-order questions, 75\% (Gemma) and 50\% (Qwen3) equal the answer to a shallower prefix.

\subsection{Evidence Used by Direct Answers}
\label{app:mech-evidence}

For correct answers, hiding the paragraph of the deciding record lowers the answer's log-probability by 16.7 nats for Gemma and 12.1 for Qwen3 (Fig.~\ref{fig:mech-diag}b). Paragraphs that support shorter prefixes matter far less: 0.5 and 0.2 nats when hidden one at a time, and 1.2 and 0.25 nats when hidden together. Hiding the same number of random paragraphs gives 0.06 and 0.02 nats. At every effective order, about one paragraph per answer passes the 1-nat threshold.

We separately examine wrong answers that equal a shallower prefix's answer (214 for Gemma, 187 for Qwen3). In 81\% and 90\% of these cases, the paragraph supporting that prefix has the largest effect, with mean drops of 12.6 and 11.2 nats. Hiding the deciding paragraph of the full query changes the log-probability by only 0.01 and 0.13 nats. For this class of errors, the answer depends mainly on evidence for a shallower query.

\subsection{Activation Patching}
\label{app:mech-patching}

\begin{figure*}[t]
\centering
\includegraphics[width=0.9\textwidth]{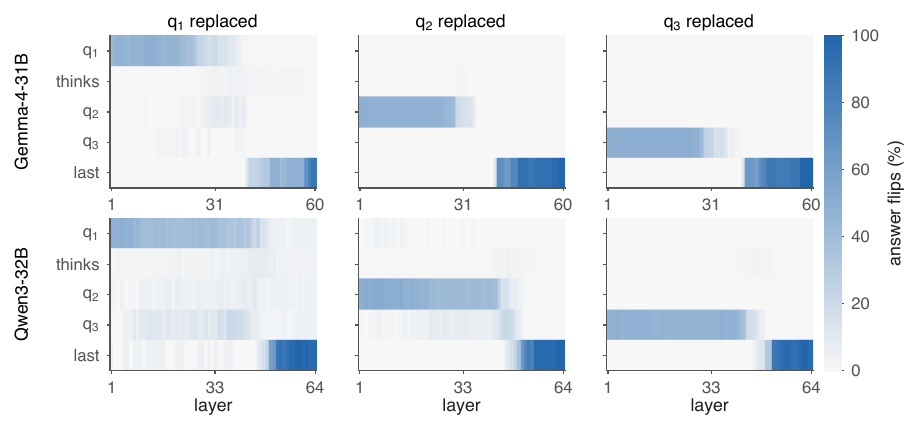}
\vspace{-0.4cm}
\caption{Token-by-layer patching on third-order pairs that differ in one agent. Each cell gives the share of pairs whose answer flips when the residual vector at that question token (rows) and layer (columns) is patched. Columns: the replaced agent $q_1$, $q_2$, or $q_3$.}
\vspace{-0.2cm}
\label{fig:mechA3}
\end{figure*}

On third-order pairs that differ in one agent, the changed agent's information moves from its name to the final token at similar depths for all three query positions (Fig.~\ref{fig:mechA3}). Patching the other agents' tokens has little effect, apart from a weak signal at later agents' tokens when $q_1$ is replaced. These interventions do not rule out intermediate nested representations.

\subsection{Multiple Choice}
\label{app:mech-mc}

\begin{figure*}[t]
\centering
\includegraphics[width=\textwidth]{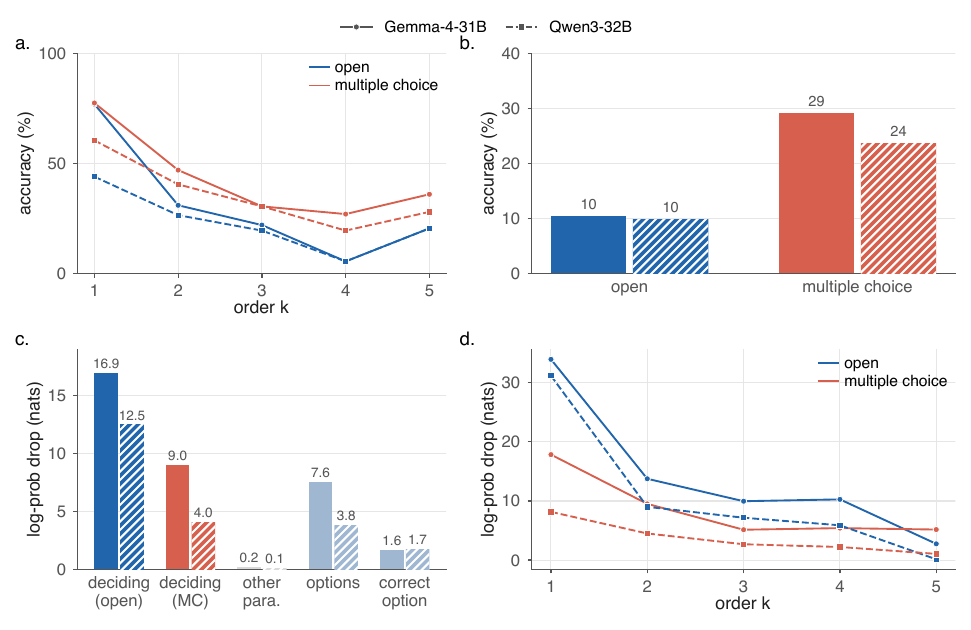}
\vspace{-0.2cm}
\caption{Open-ended and seven-option multiple-choice answers by Gemma 4 31B (solid lines and bars) and Qwen3-32B (dashed lines, hatched bars). (a) Accuracy on certified questions by order. (b) Accuracy on shortcut-resistant certified questions of order $k\geq2$. (c) Correct multiple-choice answers: log-probability drop when the deciding paragraph is hidden (open-ended and multiple-choice formats), another paragraph is hidden, the option block is hidden, or only the correct option line is hidden. (d) Drop when the deciding paragraph is hidden, by order.}
\label{fig:mechA6}
\end{figure*}

With seven options, including ``I don't know'', accuracy rises from 33\% to 45\% for Gemma and from 25\% to 37\% for Qwen3 (Fig.~\ref{fig:mechA6}a--b). Both models also favor option A, although gold positions are balanced across labels (Appendix~\ref{app:mc}).

Among correct multiple-choice answers, the deciding paragraph is still the most influential paragraph in 74--75\% of cases, including answers that are correct only with options (Fig.~\ref{fig:mechA6}c). The options add a second source of answer information. Hiding the deciding paragraph costs less than in the open-ended format: 9.0 instead of 16.9 nats for Gemma and 4.0 instead of 12.5 for Qwen3. Hiding only the correct option line costs 1.6--1.7 nats. At fifth order, the deciding paragraph's effect for Qwen3 falls to about 1 nat, so these answers rely more on the options, consistent with answering by elimination (Fig.~\ref{fig:mechA6}d).

\subsection{\ours Encoding}
\label{app:mech-hoptrace}

With one encoding call and programmatic query resolution, Gemma reaches 67.7\% and Qwen3 reaches 38.7\% on all 1,800 questions. To examine the encoding step, we append the generated records to the story and hide story paragraphs while measuring how much each record depends on them. Record order follows paragraph order in 96\% (Gemma) and 99\% (Qwen3) of cases. The record selected by the solver depends on the deciding paragraph in 89\% of correct answers but in only 13--26\% of wrong answers. Many of the remaining errors therefore originate in the encoded records.

\FloatBarrier

\end{document}